\documentclass[twoside]{article}

\usepackage{microtype}
\usepackage{graphicx}
\usepackage{mathtools}
\usepackage{amssymb}
\usepackage{amsthm}
\usepackage{bbm}
\usepackage{subcaption}
\usepackage{booktabs} 
\usepackage{tikz}
\usepackage{float}
\usepackage{ifthen}
\usepackage{pifont}
\usepackage{multirow}
\usepackage{enumitem}

\usetikzlibrary{calc,arrows.meta,positioning,snakes,fit}

\newtheorem{lemma}{Lemma}

\newtheorem{assumption}{Assumption}

\newcommand\independent{\protect\mathpalette{\protect\independenT}{\perp}}
\def\independenT#1#2{\mathrel{\rlap{$#1#2$}\mkern2mu{#1#2}}}
\usepackage[utf8]{inputenc} 
\usepackage[T1]{fontenc}    
\usepackage{hyperref}       
\usepackage{url}            
\usepackage{booktabs}       
\usepackage{amsfonts}       
\usepackage{nicefrac}       
\usepackage{microtype}      
\usepackage{xcolor}         
\usepackage{cleveref}
\usepackage[most]{tcolorbox}
\newcommand{\R}{\mathbb{R}}

\newcommand{\survS}{S(t|x)}
\newcommand{\survShat}{\hat{S}(t|x)}

\DeclareMathOperator{\sigmoid}{sigmoid}
\DeclareMathOperator{\softrelu}{SoftReLU}
\usepackage[round]{natbib}

\definecolor{olive}{rgb}{0.6, 0.6, 0.2}
\definecolor{sand}{rgb}{0.8666666666666667, 0.8, 0.4666666666666667}
\definecolor{wine}{rgb}{0.5333333333333333, 0.13333333333333333, 0.3333333333333333}
\definecolor{deblue}{RGB}{11,132,147}
\definecolor{ocra}{RGB}{204, 119, 34}

\newtcolorbox{CatchyBox}[2][]{
    lower separated=false,
    colback=white!80!sand!90!ocra,
    colframe=white, fonttitle=\bfseries,
    colbacktitle=white!50!sand!90!ocra,
    coltitle=black,
    enhanced,
    attach boxed title to top left={xshift=.02\linewidth,yshift=-4mm},
    title=#2,#1}

\usepackage[accepted]{aistats2022}
\begin{document}

\twocolumn[

\aistatstitle{Survival Regression with Proper Scoring Rules and Monotonic Neural Networks}

\aistatsauthor{ David Rindt* \And Robert Hu* \And David Steinsaltz \And Dino Sejdinovic }

\aistatsaddress{ University of Oxford } ]

\begin{abstract}
We consider frequently used scoring rules for right-censored survival regression models such as time-dependent concordance, survival-CRPS, integrated Brier score and integrated binomial log-likelihood, and prove that neither of them is a proper scoring rule. This means that the true survival distribution may be scored worse than incorrect distributions, leading to inaccurate estimation. We prove that, in contrast to these scores, the right-censored log-likelihood is a proper scoring rule, i.e., the highest expected score is achieved by the true distribution. Despite this, modern feed-forward neural-network-based survival regression models are unable to train and validate directly on the right-censored log-likelihood, due to its intractability, and resort to the aforementioned alternatives, i.e., non-proper scoring rules. We therefore propose a simple novel survival regression method capable of directly optimizing log-likelihood using a monotonic restriction on the time-dependent weights, coined SurvivalMonotonic-net (SuMo-net). SuMo-net achieves state-of-the-art log-likelihood scores across several datasets with 20--100$\times$ computational speedup on inference over existing state-of-the-art neural methods, and is readily applicable to datasets with several million observations.
\end{abstract}
\vspace{-0.2cm}
\section{INTRODUCTION AND RELATED WORKS}
\vspace{-0.1cm}

Survival analysis is a class of statistical methods for analyzing the time until the occurrence of an event, such as death in biological organisms \citep{organism_cite} or failure in mechanical systems \citep{6879441}. We focus on survival regression, where we relate the time that passes before some event occurs to one or more covariates that may be associated with that quantity of time. In particular, we consider the case in which the event times are right-censored, meaning that for some individuals it is known their event happened after a certain time, but not exactly when. For example, a patient who is still alive when the study ends would be right censored. Applications of survival regression on right-censored data are numerous, with the most classical example being the study of recovery time from medical treatments \citep{laurie1989surgical}. Other applications include modeling the time until the default of a business \citep{Dirick2017}, and churn prediction \citep{VANDENPOEL2004196}.

\textbf{Related Work}\quad Classical models for survival regression include the Cox proportional hazards model \citep{cox1972regression}, the accelerated failure time \citep{MacKenzie1982TheSA}, as well as various parametric models \citep{Kle2005}. Over the last decade, many machine learning methods for survival data have been proposed. We here mention the most relevant methods which allow more flexible survival functions than the Cox model, and refer to \citet{kvamme2019time} for further references. In \cite{katzman2018deepsurv} the Cox model is extended by replacing the linear function with a neural network. \cite{kvamme2019time} further extends on this idea and includes the survival time to the neural network, achieving state-of-the-art performance. In \cite{ishwaran2008random} a random-forest approach is taken, in which cumulative hazard functions are computed using the Nelson--Aalen estimator. In \cite{lee2018DeepHit} a discrete distribution is learned by optimizing a mix of the discrete likelihood and a rank-based score. \cite{ranganath2016deep} and \cite{miscouridou2018deep} first learn a latent representation of the covariates, and then learn a survival function using the latent representation. In particular, the latter method can deal with missing data in the context of electronic health records. In \cite{chapfuwa2018adversarial} an adversarial model is used.

\begin{figure*}
\begin{CatchyBox}{Right-censored survival regression}
	    \begin{minipage}[h]{0.34\linewidth}
	    	
		\begin{equation*}\label{eq:1}
		{
			\left\{
			\begin{aligned}
			F(t|x) &= \int_0^t f(t'|x)dt' \\
			S(t|x) &= 1- F(t|x) \\
			S_Z(t|x)&= 1-\int_0^t f(t'|x)1_{t'<Z} dt' \\
			\end{aligned}
			\right.} 
		\end{equation*}

	\end{minipage}
    \hfill
    \begin{minipage}[h]{.60\linewidth}\small
        \centering
        \begin{tabular}{r|c|c}
            Observed event-time & $t_i$ & $\R_{\geq 0}$\\\hline
            Censoring time & $c_i$ & $\R_{\geq 0}$\\\hline
            Right-censored times & $z_i=\min\{t_i,c_i\}$&$\R_{\geq 0}$ \\\hline
            Indicator of observed event-time & $d_i$ & $\{0,1\}$\\\hline
                        Covariates & $x_i$ & $\mathbb{R}^d$\\\hline
        \end{tabular}
    \end{minipage}
\end{CatchyBox}
\vspace{-0.4cm}
\end{figure*}
\textbf{Intractable Likelihood} \quad
As we model a survival \emph{distribution}, it is natural to consider approaches based on maximizing the likelihood. In the particular case of right-censoring, we later prove that the right-censored likelihood actually is a proper scoring rule \citep{gneiting-raftery07}, meaning that the highest expected score is obtained by the true distribution. However, none of the survival regression methods overviewed above have a tractable likelihood, impeding its direct use in estimation. In particular, \cite{kvamme2019time,  katzman2018deepsurv} and \cite{ishwaran2008random} all learn discrete distributions, consisting of point masses at fixed times. In \cite{lee2018DeepHit}, time is discretized into bins and while \cite{ranganath2016deep}, and \cite{miscouridou2018deep} take a Bayesian approach, the resulting posterior distribution is also intractable. In each case computing the likelihood of new observations is problematic, as no method has direct access to the cumulative survival distribution. Some recent work by \cite{groha2021general} and \cite{tang2020soden} proposes NeuralODEs to model the right-censored survival likelihood directly as a differential equation in time. While NeuralODEs are a very flexible class of models, they instead have pathologies relating to scalability \citep{dupont2019augmented}, since they scale polynomially in the number of parameters used \citep{massaroli2021dissecting}.\newline \\\textbf{Alternative scoring rules} \quad While the likelihood is a standard measure of model fit for uncensored data, due to its intractability in right-censored data one typically resorts to alternative scoring rules. The most commonly used scores to evaluate the fit of a right-censored regression model are the time-dependent concordance \citep{antolini2005time}, the survival-CRPS \citep{avati2020countdown}, the Brier score for right-censored data \citep{Graf1999} and the binomial log-likelihood. In this paper we explore pathologies of these alternative scoring rules and propose \emph{SurvivalMonotonic neural net} (SuMo-net) as a scalable and flexible survival regression model that optimizes likelihood directly. We summarize our contributions as:
\begin{enumerate}[noitemsep]
    \item We show theoretically and experimentally that time-dependent concordance, (integrated) Brier score and binomial log-likelihood, survival-CRPS are not proper scoring rules, meaning inaccurate distributions may achieve better scores than the true distribution.
    \item We give a simple proof that right-censored log-likelihood is a proper scoring rule.
    \item We introduce a novel survival regression model which directly optimizes right-censored log-likelihood and compare it to the existing methods in terms of performance and scalability, achieving state-of-the-art results on a range of datasets, together with 20--100 times speedups. Codebase: \href{https://github.com/MrHuff/Sumo-Net.git}{\texttt{https://github.com/MrHuff/Sumo-Net.git}}
\end{enumerate}
\vspace{-0.2cm}
The paper is organized as follows: In Section 2 we introduce the notation used for survival regression. In Section 3, we define what a proper score is and show that time-dependent concordance, (integrated) Brier score and binomial log-likelihood, survival-CRPS are not proper scores. We further prove that the right-censored likelihood is a proper score. \emph{SurvivalMononotic} neural net is introduced in Section 4. Section 5 provides extensive experiments. We conclude our work in Section 6.



\vspace{-0.2cm}
\section{BACKGROUND AND NOTATION} \label{sec:setting}
\vspace{-0.1cm}

We are interested in an event-time $T \in \mathbb R_{\geq 0}$ and how it depends on a covariate $X\in \mathbb R^p$ for $p\geq 1$. We consider the case in which $T$ is subject to right-censoring, where the event-time $T$ is not known for every individual in the sample. Instead of observing $T$ directly, we observe $Z = \min\{T,C\}$ where $C\in \mathbb R_{\geq 0}$ is a censoring time, as well as the indicator $D =  1 \{Z =T\}$ indicating if we observe the event-time. A sample of size $n$ can thus be denoted by $\{(X_i,Z_i,D_i) \}_{i=1}^n$, where we use the uppercase letters when we treat the dataset as random and lowercase letters otherwise. We let $S(t\vert x) \coloneqq 1 - F(t\vert x) = \mathbb P(T>t\vert X=x)$ denote the (conditional) survival distribution, and, assuming $T$ has a density, we let $f(t|x) \coloneqq \frac{d}{dt}F(t\vert x)$ denote the density of the event-time. 


The goal of this work is the following: If we let $S(t\vert x)$ denote the survival function of $T$, our aim is to estimate $S$ with $\hat S(t \vert x)$ based on the sample  $\{(X_i,Z_i,D_i) \}_{i=1}^n$. We also look at ways to assess if $\hat S$ is an accurate estimate of the true distribution $S$. 

In survival analysis one typically assumes that censoring is not informative about the event-time conditional on the covariate, which we formally state in the following assumption. 
\begin{assumption}\label{assumption:censoring} (Independent censoring) We assume that conditionally on the covariate, the censoring- and event-time are independent. That is, we assume  ${T \independent C \vert X}$. 
\end{assumption}
Let $S(t\vert x)$ be a survival function and let ${f(t\vert x)=-\partial S(t\vert x)/ \partial t}$. Then under Assumption \ref{assumption:censoring}, the right-censored log-likelihood is defined as:
\begin{equation}
    \log L_n = \sum_{i=1}^n d_i\log f(z_i\vert x_i) + (1-d_i) \log S(z_i\vert x_i).
    \label{eqn:likelihood}
\end{equation}
Both the density- and the survival function imply that, if one models the density, then the likelihood contains $n$ integrals which need to be estimated, making it a difficult optimization objective (see e.g. p. 288 of \cite{gu2013smoothing}). For that reason many methods optimize the partial likelihood instead of the right-censored likelihood.  \citep{cox1972regression}. 

\vspace{-0.2cm}
\section{EVALUATION CRITERIA AND PROPER SCORING RULES}
\vspace{-0.1cm}

A scoring rule $\mathcal S$ takes a distribution $S$ over some set $\mathcal Y$ and an observed value $y\in \mathcal Y$ and returns a score $\mathcal S( S, y)$, assessing how well the model predicts the observed value. For a positive scoring rule, a higher score indicates a better model fit. A scoring rule is called proper if the true distribution achieves the optimal expected score, i.e., if
\begin{align*}
    \mathbb E_{y\sim S}\mathcal S(S,y) \geq \mathbb E_{y \sim S}\mathcal S(\hat S, y) \quad \text{for all distributions }  \hat S.
\end{align*}
 In the context of survival regression, we call the scoring rule $\mathcal S$ proper if for every true distribution $S$, every censoring distribution $C$ and for every covariate $x$ it holds that
\begin{align*} \label{eqn:propers_score_survival_regression}
    \mathbb E_{T,C\vert X=x} \mathcal S \bigl( \survS, (Z, D) \bigr) \geq \mathbb E_{T,C\vert X=x} \mathcal S\bigl( \survShat, (Z, D)\bigr)
\end{align*}
for every distribution $\survShat$, where in the above expression $T\vert X=x \sim \survS$. By taking the expectation with respect to $X$ in the above expression, we find that if $\mathcal S$ is a proper score, then also 
\begin{align*}
    \mathbb E_{T,C,X} \mathcal S\bigl( {S}(t|X), (Z, D) \bigr) \geq \mathbb E_{T,C, X} \mathcal S\bigl( \hat{S}(t|X), (Z, D) \bigr)
\end{align*}
for every family of distributions $\survShat$ with $x\in \mathbb R^p$. 

\subsection{Existing evaluation metrics are not proper scoring rules}\label{sec:other_score_not_proper}

We now show that time-dependent concordance \citep{antolini2005time}, the Brier score for right-censored data \citep{Graf1999}, the binomial log-likelihood and the survival-CRPS \citep{avati2020countdown} are not proper scoring rules. For time-dependent concordance and the survival-CRPS, these are a novel results. Time-dependent concordance is not to be confused with \citep{10.1093/biostatistics/kxy006}, which considers \emph{time-independent} concordance.  

\begin{figure*}
\centering
\begin{subfigure}{.5\textwidth}
  \centering
  \includegraphics[width=1\linewidth]{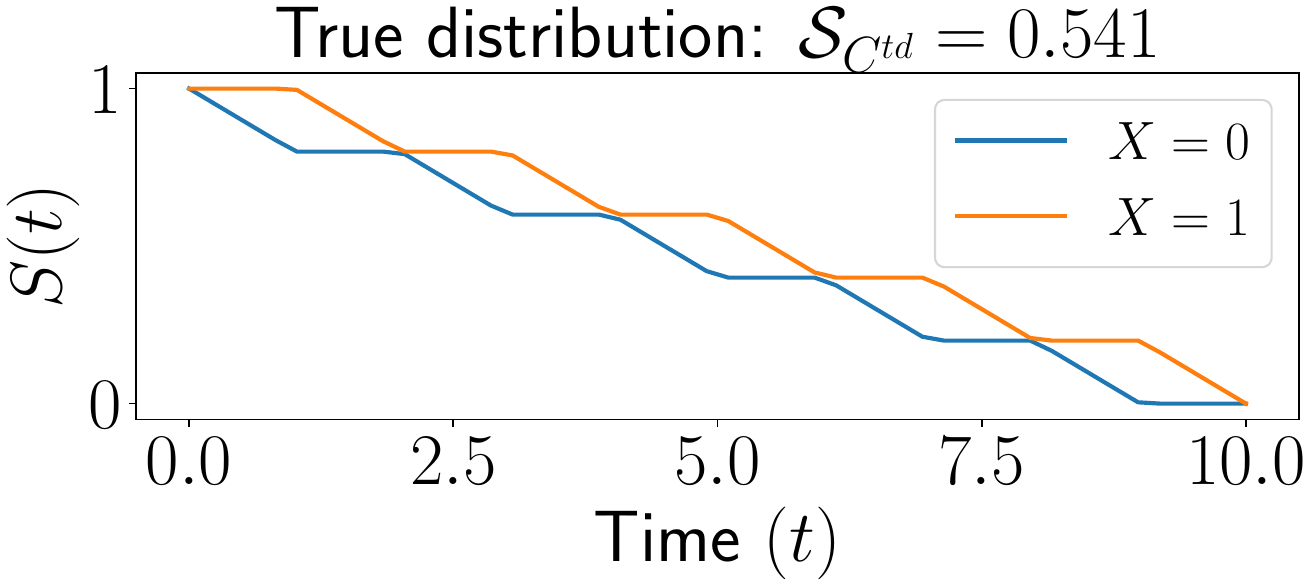}
\end{subfigure}%
\begin{subfigure}{.5\textwidth}
  \centering
  \includegraphics[width=1\linewidth]{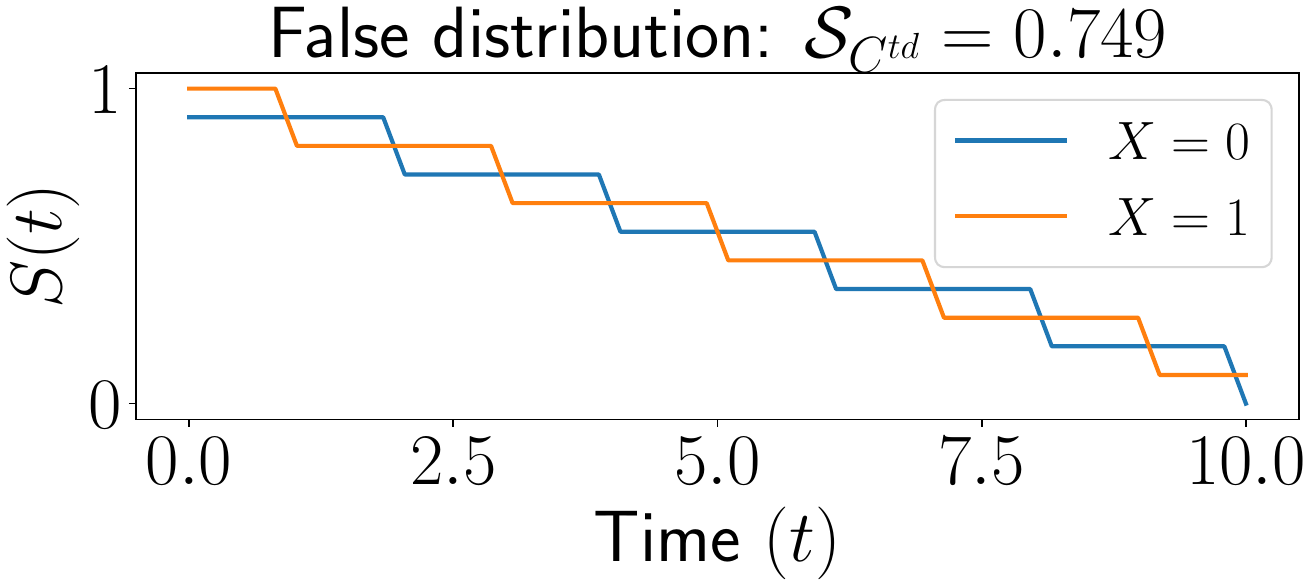}
\end{subfigure}%
\vspace{-0.2cm}
\caption{\label{fig:concordance_example} The survival curves for the true distribution on the left and a false distribution, designed to optimize concordance, on the right. By drawing a sample of size $1000$ from the true distribution, we find that the concordance score of the true distribution is lower than the score of the false distribution, confirming that an inaccurate distribution can have high concordance.  }
\vspace{-0.4cm}
\end{figure*}

\textbf{Time-dependent concordance} \quad Let $(X,Z,D)$ and $(X',Z',D')$ be i.i.d.\ covariates and right-censored event-times. Time-dependent concordance $\mathcal{S}_{C^{\text{td}}}$ \citep{antolini2005time} is defined by
    \begin{align*}
        \mathbb P\left( \hat S(Z\vert X) < \hat S(Z \vert X') \vert Z \leq Z', D =1\right)\\+\frac12 \mathbb P\left( \hat S(Z\vert X) = \hat S(Z \vert X') \vert Z \leq Z', D =1\right)
    \end{align*}
where ties in the survival probabilities are dealt with as proposed in \cite{ishwaran2008random} (Step 3 of Section 5.1). It has been remarked that poorly calibrated models can still have a high time-dependent concordance \citep{kvamme2019time}. \cite{rizopoulos2014combining} and \cite{antolini2005time} have pointed out that censoring affects time-dependent concordance. We now show that time-dependent concordance is not a proper score by constructing an example where a false distribution has higher score than the true distribution. \\\textit{Example: optimizing concordance for a binary covariate.} Let $X, X'\sim \text{Bernoulli}(1/2)$ independently. Then $\hat S(\cdot\vert \cdot)$ maximizes $\mathcal{S}_{C^{\text{td}}}$ if and only if it maximizes 
\begin{align*}
    & \mathbb P \bigl(\hat S(Z\vert X)\!<\!\hat S(Z\vert X') , Z \!\leq\! Z', D\!=\!1 \vert X \!= \!1, X'\!=\!0 \bigr) +\\ 
    & \mathbb P\bigl(\hat S(Z\vert X)\!<\!\hat S(Z\vert X') , Z \!\leq\! Z', D\!=\!1 \vert X \!= \!0, X'\!=\!1 \bigr)\!=
    \\   & \int_0^{\infty}  \big( 1\{\hat S(t\vert1)\!<\!\hat S(t\vert0) \} f_T(t\vert 1)S_C(t\vert 1)S_T(t\vert 0)S_C(t\vert0)  
    \\   &+  1\{\hat S(t\vert 0)\!<\!\hat S(t\vert 1) \} f_T(t\vert 0)S_C(t\vert 0)S_T(t\vert 1)S_C(t\vert 1) \big) dt.
\end{align*}
This expression is optimized by constructing $\hat S(\cdot \vert \cdot)$ such that $\hat S(t\vert 1) < \hat S(t\vert 0)$ if and only if 
\begin{align*}
         f_T(t\vert 1)S_C(t\vert 1)S_T(t\vert 0)S_C(t\vert0) &\geq
         \\ f_T(t\vert 0)S_C(t\vert 0)S_T(t\vert 1)S_C(t\vert 1) 
         \iff \lambda_T(t\vert 1) & \geq \lambda_T(t\vert 0)
\end{align*}
where $\lambda_T(t\vert x) = f_T(t\vert x)/S_T(t\vert x)$ is called the hazard rate. Hence, in this example, to maximize concordance we need the ordering of $\hat S(t \vert 1)$ and $\hat S(t\vert 0)$ to be the reversed order of the true hazard rates. In Figure \ref{fig:concordance_example}, we use this insight to generate a false distribution with concordance much higher than the concordance of the true distribution (Example 1). See Appendix 1.2 for details.

\begin{table*}
\resizebox{\linewidth}{!}{%


\begin{tabular}{|l|cc|ccccc|cc|}
\hline
         & \multicolumn{2}{c|}{Example 1}                                     & \multicolumn{5}{c|}{Example 2}                                                                                                                                                  & \multicolumn{2}{c|}{Example 3}                                     \\
         & $\mathcal{S}_{L}\uparrow$ & $\mathcal{S}_{C^{\text{td}}}\uparrow$ & $\mathcal{S}_{L}\uparrow$ & $\mathcal S_{\text{Brier}}^{t=4.0}\downarrow$ & $\mathcal S_{\text{BLL}}^{t=4.0}\uparrow$ & $\mathcal S_{\text{IBS}}\downarrow$ & $\mathcal S_{\text{IBLL}}\uparrow$ & $\mathcal{S}_{L}\uparrow$ & $\mathcal{S}_{\text{CRPS}}\downarrow$ \\ \hline
True distribution score     & $-1.375$                  & $0.541$                               & $-1.795$                  & $0.220$                             & $0.085$                           & $0.032$                             & $-0.009$                           & $-0.513$                  & $3.884$                               \\
Fake distribution score     & $-11.029$                 & $0.749$                               & $-1.820$                  & $0.204$                             & $0.122$                           & $0.031$                             & $-0.006$                           & $-0.626$                  & $1.646$                               \\
Consistent? & $\checkmark$              & \ding{53}           & $\checkmark$              & \ding{53}         & \ding{53}       & \ding{53}         & \ding{53}        & $\checkmark$              & \ding{53}           \\ \hline
\end{tabular}
}
\vspace{-0.2cm}
\caption{Scores for true and fake distribution respectively. Bottom line indicates if the true distribution scores better than the fake distribution. Right-censored likelihood consistently scores the true distribution higher than a fake distribution, in contrast to non-proper scores.}
\label{breaking_non_proper_scores}
\vspace{-0.4cm}
\end{table*}

\textbf{Brier score} \quad The Brier score at time $t$ for the estimated survival distribution $\hat S(\cdot \vert \cdot)$ is typically defined as \citep{Graf1999}
\begin{align*}
    \mathcal S^t_{\text{Brier}}&(\survShat, (z,d))= \\& \frac{\survShat^21\{z \leq t, d=1 \}} {\hat G(z)}  + \frac{( 1 -\survShat)^21\{z>t \} }{\hat G(t)}
\end{align*}
where $\hat G$ is the Kaplan-Meier estimate of the censoring distribution. Assuming that the estimated survival function of the censoring time $C$, $\hat G(\cdot)$ is the exact marginal distribution $G(\cdot) = \mathbb P(C>\cdot)$, then the expectation, conditional on the $X_i$ equals
\begin{align}
     \mathbb  E\bigl[  \mathcal S_{\text{Brier}}&(\survShat, (Z,D))\vert X=x\bigr] = \\ &  \label{eqn:expected_brier_score} \survShat^2 \int_0^t   \frac{G(z\vert x)} { G(z)}   f_T(z\vert x)dz \\&+  ( 1 -\survShat)^2 \frac{G(t\vert x)}{G(t)}S_T(t\vert x).
\end{align}
If $C\independent X$, and $G(\cdot \vert x) = G(\cdot)$, for all $x$, then it is easy to see the above equals
\begin{align*}
 \survShat^2  (1-S_T(t\vert x)) + (1-\survShat)^2 S_T(t\vert x).
\end{align*}
By setting $a_i=\survShat$ and optimizing with respect to each of the $a_i$, we find that the expected Brier score is minimized for $\survShat=S_T(t\vert x)$, i.e., when the estimated survival probabilities equal the true survival probabilities. \emph{Under the assumption of censoring independent of the covariate and a perfectly estimated censoring distribution, } the Brier score is thus a  proper score. However, as Equation \ref{eqn:expected_brier_score} makes clear, the Brier score may not be a proper score when $C\not \independent X$. The problematic assumption on the censoring distribution has, for example, been reported in \cite{kvamme2019brier}. We construct an example (Example 2) where an inaccurate distribution achieves a lower (i.e., better) Brier score than the true distribution in \Cref{breaking_non_proper_scores}. For details see Appendix 1.3. Indeed, one could use estimates that include the covariate instead of the Kaplan--Meier estimate, so as to estimate $\mathbb{P}(C > t\vert x)$. Since this requires a regression of the censoring distribution to compute a score of the regression of the event-time, this raises the question of how to choose the best regression of the censoring distribution. Because of this loop, in practice the Kaplan--Meier estimate is often chosen \citep{kvamme2019brier, kvamme2019time}. The same analysis of the Brier score can be applied to the binomial log-likelihood (BLL). In particular, the BLL is a proper score under the assumption that $C\independent X$, but it need not be proper when this assumption is violated. We use Example 2 to demonstrate that BLL also breaks in \Cref{breaking_non_proper_scores}. It should be stressed that the pathology of Brier Score lies in the difficulty of estimating the weights rather than the score itself; work such as \cite{han2021inverseweighted}, propose inverse weighted survival games to correctly estimate the Brier Score. \newline \\ \textbf{Integrated Brier score and integrated binomial log-likelihood} \quad The Brier score above was defined for a fixed time $t$. The integrated Brier score (IBS) has been defined as a measure of goodness-of-fit of the estimate $\hat S(\cdot\vert \cdot)$ by
\begin{align*}
    \mathcal S_{\text{IBS}}& \bigl(\survShat, (z,d) \bigr) \\&= \frac{1}{t_2-t_1}\int_{t_1}^{t_2} \mathcal S^t_{\text{Brier}}\bigl(\survShat, (z,d) \bigr) dt,
\end{align*}
where typically $t_1=0$ and $t_2=z_{\text{max}}=\max \{z_1,\dots,z_n\}$. In experiments, the integral is approximated by a sum over an equally spaced grid of times. The binomial log-likelihood at time $t$, $\mathcal{S}_\text{BLL}^t$ is defined by: 
\begin{align*}
    \mathcal S_{\text{BLL}}^t & \bigl(\survS, (z,d) \bigr) = \\& \frac{\log(1-\survShat)1\{z\leq t, d=1\}}{\hat G(z)}  + \frac{ \survS1\{z\geq t\}}{\hat G(t)}.
\end{align*}
Analogously to the definition of IBS, the integrated binomial log-likelihood, IBLL, is defined as
\begin{align*}
    \mathcal S_{\text{IBLL}}(\survS,(z,d))= \frac{1}{t_2-t_1}\int_{t_1}^{t_2} \mathcal S^t_{\text{BLL}}(\survS,(z,d))dt.
\end{align*}
As neither $\mathcal{S}_{\text{Brier}}^t$ nor $S^t_{\text{BLL}}$ are proper scores, it follows that $S_{\text{Integrated-Brier}}$ and $\mathcal S_{\text{IBLL}}$ are not proper scores. We use Example 2 to show that IBS and IBLL breaks in \Cref{breaking_non_proper_scores}.


\textbf{Survival-CRPS} \quad The survival continuous ranked probability score (survival-CRPS) is defined by
\begin{align*}
    &\mathcal S_{\text{CRPS}}\bigl(\hat F_X, (Z,D)\bigr)  =\\& \int_0^Z \hat F(t \vert X)^2dt + D \int_Z^{\infty} (1-\hat F(t\vert X))^2 dt
\end{align*}
    and was proposed in \cite{avati2020countdown} as an extension of the continuous ranked probability score (CRPS) \citep{gneiting2007probabilistic}. This is equivalent to the integrated Brier score without inverse probability weighting. In \cite{avati2020countdown}, the score was also claimed to be a proper score. The proof strategy given there argues that the survival-CRPS is simply the weighted CRPS of \cite{gneiting2011comparing} with the weight function set to the indicator of uncensored regions. As we show in Appendix 1.4.1, however, this weight function does not recover the survival-CRPS. This suggests that the survival-CRPS is not a proper score. For example, if the censoring time $C$ is deterministic, i.e., $C=c$ w.p. 1 for $c\in \mathbb R_{\geq 0}$, then it is clear that to minimize the survival-CRPS, one would set $\hat F(c)=1$, even if the true CDF is less than 1. This principle also holds in the case of random censoring, where one can construct false CDFs which yield a lower (i.e., better) survival-CRPS. We use this insight to construct an example (Example 3) where survival-CRPS breaks in \Cref{breaking_non_proper_scores}. For details, see Appendix 1.4.2.
    

\subsection{The right-censored log-likelihood is a proper score}\label{sec:log_likelihood_proper_score}

The right-censored log-likelihood score of an observation $(z,d)$, given a covariate $x$ and distribution $\survS$ is simply the likelihood given in Equation \ref{eqn:likelihood}, i.e.: 
\begin{align*}
    \mathcal S_{L}(\survS, (z,d)) = d \log f(z\vert x) + (1-d) \log S(z\vert x).
\end{align*} 
Let $\hat S$ denote any continuous survival curve, and let $S$ denote the true survival distribution of the event-time $T$, and make the following assumption.  
\begin{assumption} \label{assumption:kl_divergence}
Assume that $\text{KL}(S \vert \vert \hat S)<\infty$, where $\text{KL}$ denotes the Kullback--Leibler divergence. 
\end{assumption}
Under this assumption, we prove the following lemma.
\begin{lemma}
\label{lemma:likelihood_proper_score}
Let $S$ be the true survival distribution of the event-time $T$. For every distribution $\hat S$ that satisfies Assumption \ref{assumption:kl_divergence} and for every $x\in \mathbb R^p$, it holds that
\begin{align*}
           \mathbb E \left[ \mathcal S _{L}(\survS, (Z,D)) \vert X=x\right] \\ \geq \mathbb E [  \mathcal S _{L}(\survShat, (Z,D)) \vert X=x ]
\end{align*}
\end{lemma}
\begin{proof}
We need to prove that for every $x\in \mathbb R^p$, for every distribution $\survS$
\begin{align*}
  \mathbb E \left[ \mathcal S (\survS, (Z,D)) \vert X=x\right] \\\geq \mathbb E [  \mathcal S (\survShat, (Z,D)) \vert X=x ]
\end{align*}
where $T \sim \survS$, the censoring time $C$ follows an arbitrary distribution and it is assumed that $\survS$ and $\survShat$ satisfy Assumption 1. We can safely omit $x$ from the notation and prove that 
\begin{align*}
  \mathbb E \mathcal S (S, (Z,D))   \geq \mathbb E \mathcal S (\hat S, (Z,D))
\end{align*}
where again $T\sim S$, the censoring time $C$ follows an arbitrary distribution and it is assumed that $S$ and $\hat S$ satisfy Assumption 1. This has been proved in a slightly different context in \cite{diks2011likelihood} Lemma 1. In that context the censoring variable is assumed known. The same argument can be applied to the case of random censoring, by first conditioning on the censoring variable. Namely, if we can prove that
\begin{align*}
    \mathbb E  \left[ \mathcal S( S, (Z,D)) \vert C \right] \geq \mathbb E [ \mathcal S(\hat S, (Z,D)) \vert C ]
\end{align*}
then it follows, by taking the expectation over $C$, that also 
\begin{align*}
        \mathbb E \mathcal S( S, (Z,D) ) \geq \mathbb E \mathcal S(\hat S, (Z,D) ).
\end{align*}
We now prove the first inequality following the proof  of Lemma 1 in  \cite{diks2011likelihood}. Let $f$ denote the density of $S$, and let $\hat f$ denote the density of $\hat S$. Note that the expected score can be written as follows: 
\begingroup
\allowdisplaybreaks
\begin{align*}
  &\mathbb E \left[\mathcal S(\hat S, (Z,D))\vert C\right]=  \\& \mathbb E \left[  D \log \hat f(Z) +  (1-D) \log\hat S(Z)  \big \vert C \right]
  \\ & =\int_0^{\infty} 1\{t\leq C \}f(t)\log \hat f(t) + 1\{t > C \}f(t)\log \hat S(C) dt 
  \\ & = \int_0^{C} f(t) \log \hat f(t) dt + S(C)\log \hat S(C).
\end{align*}
\endgroup
Hence, 
\begingroup
\allowdisplaybreaks
    \begin{align*}
  &\mathbb E \bigl[ \mathcal S(S, (Z,  D)) - \mathcal S(\hat S, (Z,D)) \,\bigm| \, C \bigr] \\ &=\int_0^{C} f(t) \log \frac{f(t)}{\hat f(t)} dt + S(C)\log \frac{ S(C)}{\hat S(C)}\\&= F(C) \int_0^{C} \frac{f(t)}{F(C)} \log \frac{f(t)/F(C)}{\hat f(t)/\hat F(C)} dt +S(C)\log \frac{ S(C)}{\hat S(C)} -\\ &F(C) \int_0^{C} \frac{f(t)}{F(C)} \log\frac{1/F(C)}{1/\hat F(C)} dt
     \\ &= F(C) \int_0^{C} \frac{f(t)}{F(C)} \log \frac{f(t)/F(C)}{\hat f(t)/\hat F(C)} dt+\\&S(C)\log \frac{ S(C)}{\hat S(C)}+ F(C)  \log \frac{F(C)}{\hat F(C)} \\&= F(C)\text{KL}\bigl(f(t)/F(C) \,\|\,  \hat f(t) / \hat F(C) \bigr)+\\& \text{KL}\bigl(\text{Ber}(S(C)) \,\|\, \text{Ber}(\hat S(C)) \bigr)
    \geq 0.
    \end{align*}
\endgroup

The distributions $f(t)/F(C)$ and $\hat f(t) / \hat F(C)$ are the densities of $f$ and $\hat f$ conditioned on the event $T\leq C$. The first KL divergence exists by Assumption 1. The second term is the KL divergence between Bernoulli random variables with success probabilities $S(C)$ and $\hat S(C)$ respectively. 
\end{proof}
Hence, under Assumption \ref{assumption:kl_divergence}, the right-censored log-likelihood is a proper scoring rule. We provide a proof in Appendix 1.1, as a simple extension of Lemma 1 of \cite{diks2011likelihood}, by conditioning on the censoring variable and the covariate. We note that this result may also be deduced from Section 3.5 of \cite{dawid2014theory}, as a non-trivial special case where one uses the function $\psi(p) = p\log p - p$, but the more general claim there is provided without a proof. We empirically validate that the right-censored log-likelihood is a proper score by applying it to the true and fake distributions for Example 1, 2 and 3 in \Cref{breaking_non_proper_scores}. As expected, the right-censored log-likelihood correctly gives the true distribution a higher score than the fake distribution for all examples.

\textbf{Summary of insights} \quad We have shown both theoretically and empirically that $\mathcal{S}_C^{\text{td}}$, $\mathcal{S}_{\text{IBS}}$, $\mathcal{S}_{\text{IBLL}}$ and $\mathcal{S}_{\text{CRPS}}$ are not proper scores, meaning that optimizing them gives no guarantee of learning the true survival distribution for right-censored data. Since the right-censored log-likelihood is proven to be a proper score, it is imperative to construct models that train and evaluate on the log-likelihood directly, rather than using non-proper proxy scores.

\vspace{-0.2cm}
\section{OPTIMIZING LIKELIHOOD USING PARTIALLY MONOTONIC NEURAL NETWORKS}
\vspace{-0.1cm}

\begin{figure*}
\centering
\scalebox{.8}{\def\layersep{1.5}
\def\hlayersep{1}

\begin{tikzpicture}[draw=black!50, node distance=\layersep]
            
    \tikzstyle{every pin edge}=[<-,shorten <=1pt]
    \tikzstyle{positive}=[snake=coil,segment aspect=0,segment amplitude=1pt];
    \tikzstyle{neuron}=[circle,fill=gray!30,minimum size=12pt,inner sep=0pt]
    \tikzstyle{input x neuron}=[neuron, fill=gray!30,text width=4mm,align=center];
    \tikzstyle{input y neuron}=[neuron, fill=gray!30,text width=4mm,align=center];
    \tikzstyle{cdf neuron}=[neuron, fill=gray!30];
    \tikzstyle{pdf neuron}=[neuron, fill=gray!30,text width=8mm,align=center];
    \tikzstyle{hidden neuron}=[neuron, fill=gray!30,text width=4mm,align=center];
    \tikzstyle{annot} = [text width=4em, text centered]
    \tikzstyle{arrow}=[shorten >=1pt,->]

    \node[input x neuron] (x1) at (0,1) {$x_1$};
    \node[input x neuron] (xd) at (0, 1+\layersep) {$x_d$};    
    \node at ($(x1)!.6!(xd)$) {\vdots};
    
    \node[hidden neuron] (xh1) at (\hlayersep, 1) {};
    \node[hidden neuron] (xhd) at (\hlayersep,1+\layersep) {};
    \node (xh_dots) at ($(xh1)!.6!(xhd)$) {\vdots};
    
    \path[arrow] (x1) edge (xh1);
    \path[arrow] (x1) edge (xhd);
    \path[arrow] (xd) edge (xh1);
    \path[arrow] (xd) edge (xhd);
    
    \node[hidden neuron] (xh_out1) at (2*\hlayersep, 1) {};
    \node[hidden neuron] (xh_outd) at (2*\hlayersep, 1+\layersep) {};
    \node (xh_out_dots) at ($(xh_out1)!.6!(xh_outd)$) {\vdots};
    
    \node at ($(xh_dots)!.5!(xh_out_dots)$) {\vdots};
    
    \node[input y neuron] (y) at (2*\hlayersep,0.5+2*\layersep) {$t$};
    
    \node[hidden neuron] (xy_1) at (3*\hlayersep, 1+0.5) {};
    	\node[hidden neuron] (xy_d) at (3*\hlayersep, 1+1.5) {};
    	\node (xy_dots) at ($(xy_1)!.6!(xy_d)$) {\vdots};
    	
    \path[arrow] (xh_out1) edge (xy_1);
    \path[arrow] (xh_out1) edge (xy_d);
    \path[arrow] (xh_outd) edge (xy_1);
    \path[arrow] (xh_outd) edge (xy_d);
    \draw[arrow,positive]  (y) -- (xy_1);
    \draw[arrow,positive] (y) -- (xy_d);
    
    \node[hidden neuron] (xy_h1_1) at (4*\hlayersep, 1+0.5) {};
    	\node[hidden neuron] (xy_h1_d) at (4*\hlayersep,1+ 1.5) {};
    	\node (xy_h1_dots) at ($(xy_h1_1)!.6!(xy_h1_d)$) {\vdots};
    	
    \draw[positive,arrow] (xy_1) -- (xy_h1_1);
    \draw[positive,arrow] (xy_1) -- (xy_h1_d);
    \draw[positive,arrow] (xy_d) -- (xy_h1_1);
    \draw[positive,arrow] (xy_d) -- (xy_h1_d);
    	
    \node[hidden neuron] (xy_h2_1) at (5*\hlayersep, 1+0.5) {};
    	\node[hidden neuron] (xy_h2_d) at (5*\hlayersep,1+ 1.5) {};
    	\node (xy_h2_dots) at ($(xy_h2_1)!.6!(xy_h2_d)$) {\vdots};
     
    \node at ($(xy_h1_dots)!.5!(xy_h2_dots)$) {\vdots};
    
    \node[cdf neuron] (h) at (6.2*\hlayersep, 1+1) {$h(t,x)$};
    
    \draw[positive,arrow] (xy_h2_1) -- (h);
    \draw[positive,arrow] (xy_h2_d) -- (h);
    
    \node[cdf neuron] (cdf) at (8*\hlayersep, 1+1) {$\hat S(t\vert x)$};
    
    \draw[snake=triangles,segment object length=3pt, segment length=3pt] (h) -- node[below]{$1-\sigma$} (cdf);

    \node[cdf neuron] (pdf) at (9.8*\hlayersep, 1+1) {$\hat f(t\vert x)$};    	
    	
    	 \draw[snake=triangles,segment object length=3pt, segment length=3pt] (cdf) -- node[below]{$-\frac{\partial}{\partial t}$} (pdf);

    	\path[arrow] (11, 2.4) edge node[at end, label=right:{$w_{ij} \in \mathbb{R}$}]{} (12, 2.4);
    	\draw[positive] (11, 3) -- node[at end, label=right:{$w_{ij} \in \mathbb{R}^+$}]{} (12, 3);
    	\draw[snake=triangles,segment object length=3pt, segment length=3pt] (11,3.6) -- node[at end, label=right:{no weights}]{} (12,3.6);
    	\node[input x neuron] (input_x_legend) at (11,0) {$x_j$} node[right=0pt of input_x_legend] {covariate};
    	\node[input y neuron] (input_y_legend) at (11,.6) {$t$} node[right=0pt of input_y_legend] {time}; 
    	\node[cdf neuron] (cdf_legend) at (11,1.2) {$h$} node[right=0pt of cdf_legend] {$bias+\sum_i w_{ij}input_i$};
    	\node[hidden neuron] (hidden_legend) at (11,1.8) {} node[right=0pt of hidden_legend] {$\tanh(bias+\sum_i w_{ij}input_i)$}; 
    	
    	
    	

\end{tikzpicture}}
\vspace{-0.2cm}
\caption{[Figure adapted from Figure 1 of \cite{chilinski2020neural}] The graph representing SuMo-net, modelling the survival curve. The last edge symbolizes the operation of differentiating the conditional survival distribution function $\hat S(t|x)$ with respect to the input $t$, resulting in the conditional density of the survival time. The second-to-last edge represents passing $h(t,x)$ through $1-\sigmoid$. If one models the cumulative hazard instead, we replace $1-\sigmoid$ by $\softrelu$ and use $\frac{\partial}{\partial t}$ in the last edge instead of $-\frac{\partial}{\partial t}$.}
\vspace{-0.4cm}
  \label{fig:network_structure}
\end{figure*}
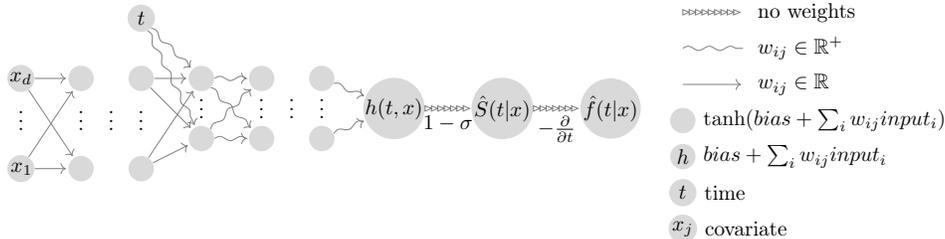

To optimize the right-censored log-likelihood of \eqref{eqn:likelihood}, one needs to jointly model the survival distribution and the density. To do so, we adapt the monotonic neural density estimator (MONDE) of \cite{chilinski2020neural} in the case of a univariate response variable to the survival context. A similar strategy has been used in \cite{10.5555/3454287.3454477} to model temporal point processes. We adapt MONDE to jointly model the survival function and the density function of a right-censored event-time, resulting in a scalable and flexible survival regression model that can train and evaluate directly on log-likelihood. The model is coined \emph{SurvivalMonotonic-neural network} (SuMo-net) and we provide a brief overview of its architecture in Figure \ref{fig:network_structure}.

\textbf{Architecture} \quad Figure \ref{fig:network_structure} pictures the SuMo-network structure as a directed graph with two types of edges. The network can be defined in terms of two fully connected feed-forward sub-networks. A covariate $x$ is first passed through a network $h^{\text{cov}}(x)$ with $L^{\text{cov}}$ layers in which each layer is fully connected to the next, and no other edges are present. The output of the covariate network, together with the time input $t$ forms the input of a second network $h^{\text{mixed}}(t,u)$, where $u$ has the same dimension as the output of the covariate network $h^{\text{cov}}(x)$. In this network, too, all edges are present from layer $l$ to $l+1$ for $l=1,\dots,L^{\text{mixed}}-1$ and no additional edges are added. The output layer of the mixed network consists of a single neuron with value $h(t,x)$, thus returning a scalar. The total network $h$ thus equals 
\begin{equation*}
    h(t,x) = h^{\text{mixed}}(t, h^{\text{cov}}(x)).
\end{equation*}
We then model $S(t\vert x)$ through $\hat S(t\vert x) = 1-\sigmoid\left(h(t,x) \right)$. It should be noted that we propose a minimal implementation of SuMo-net, only using a vanilla feed-forward architecture. Any other type of architecture such as Resnet \citep{he2015deep}, LSTM \citep{LSTM_cite}, Densenet \citep{huang2018densely} etc., can be used to enhance model flexibility further, depending on the nature of the application.  

\textbf{Training} \quad After passing $h(t,x)$ through the $\sigmoid$ function, the final edge in Figure \ref{fig:network_structure} represents differentiation of $\sigmoid(h(t,x))$ with respect to $t$ and we set $\hat f(t\vert x) = -\partial \hat S(t\vert x)/ \partial t$. To compute the derivative, we use automatic differentiation as implemented in PyTorch \citep{NEURIPS2019_9015}, in line with the approach taken in \cite{chilinski2020neural}. One can also approximate the derivative by applying the chain rule to $\sigmoid(h(t,x))$, yielding that
\begin{align*}
f(t\vert x) =& -\sigmoid(h(t,x))(1-\sigmoid(h(t,x)) \\ &\times  \left( \frac{h(t+\epsilon, x)- h(t,x)}{\epsilon}\right) + o(1).
\end{align*}
We implemented SuMo-net both using this approximation and using auto-grad and found the implementations performed comparably during the experiments. We found further that both of these implementations improved performance over simply evaluating $-\left(S(t+ \epsilon \vert x)-S(t\vert x)\right)/\epsilon$, which may be due to the vanishing gradient of the $\sigmoid$ function as one moves away from $0$.


\textbf{Universality} \quad Since the survival function and the cumulative hazard function are non-increasing and non-decreasing in $t$ respectively, we can ensure that the output of the neural net does not decrease as $t$ increases by restricting the weights which are applied to the time $t$ to be non-negative. As discussed in \cite{lang2005monotonic}, non-decreasing behaviour in $t$ is guaranteed if all weights that are descendants of $t$ in Figure \ref{fig:network_structure} are non-negative. In \cite{lang2005monotonic} it is furthermore shown that if there are at least 2 hidden layers in the mixed network $h_{\text{mixed}}(t,u)$, then, given enough nodes in each layer, the network can universally approximate any true survival function. In our implementation, we ensure that weights are non-negative by defining them as the square of unrestricted weights. It should be noted that despite being a relatively simple idea, the neural network property of SuMo-net allows survival regression to straightforwardly use \emph{any type of data} such as images, graphs or any type of sequences, which allows for an unrivaled flexibility in data choices.   
\vspace{-0.2cm}
\section{EXPERIMENTS}\label{sec:experiments}
\vspace{-0.1cm}
We apply SuMo-net to five real datasets SUPPORT, METABRIC, GBSG, FLCHAIN, and KKBOX presented in \cite{kvamme2019time}. We compare the results with the methods DeepHit \citep{lee2018DeepHit}, Cox-Time and Cox-CC \citep{kvamme2019time}, DeepSurv \citep{katzman2018deepsurv} and the classical Cox model \citep{cox1972regression}. To nuance the comparisons, we contrast against reported results in SODEN \cite{tang2020soden} and SurvNODE \cite{groha2021general}, which are both NeuralODE based models that can train on likelihood directly. We also compare against modelling the parameters (scale \& shape and mean \& variance) of the Weibull and Log-normal distribution with neural nets as functions of covariates and time, analogous to \cite{avati2020countdown}. When parametrising a survival distribution directly, the right-censored likelihood can in fact be calculated directly.

We use a 5-fold cross-validation procedure in which the model is trained on three folds, one fold is used for validation, and one fold is reserved for testing. We then average the evaluation criteria calculated on five different test folds. For datasets FLCHAIN, GBSG, METABRIC, and SUPPORT we optimize SuMo-net and baselines by using the hyperopt library \citep{10.5555/3042817.3042832} for 300 iterations. For the KKBOX dataset we do the same, but for 100 hyperopt iterations, due to computational constraints. All experiments were run on a GPU cluster consisting of 8 Nvidia 1080 GTX TI cards. 

\begin{table*}

    \centering
    \resizebox{\linewidth}{!}{%
\begin{tabular}{lllllllll}
\hline
Dataset          & SUPPORT & METABRIC & GBSG & FLCHAIN & KKBOX     & Weibull & Checkerboard & Normal \\ \hline
Size             & 8,873   & 1,904    & 2,232                      & 6,524   & 2,646,746 & 25,000  & 25,000       & 25,000 \\
Covariates       & 14      & 9        & 7                          & 8       & 15        & 1       & 1            & 1      \\
Unique Durations & 1,714   & 1,686    & 1,230                      & 2,715   & 2,344,004 & 25000   & 25000        & 25000  \\
Prop. Censored   & 0.32    & 0.42     & 0.43                       & 0.70    & 0.28      & 0.55    & 0.75         & 0.50   \\ \hline
\end{tabular}

    }
    \vspace{-0.2cm}
    \caption{Summary of all datasets used in the experiments. Weibull, checkerboard and normal are synthetic datasets and the others are real datasets. See \cite{kvamme2019time} for more details on the real datasets.}\label{tab:dataset_statistics}
    \vspace{-0.4cm}
\end{table*}

\begin{figure*}
    \centering
    \begin{subfigure}[t]{\textwidth}
        \begin{subfigure}[t]{0.33\textwidth}
        \includegraphics[width=\linewidth]{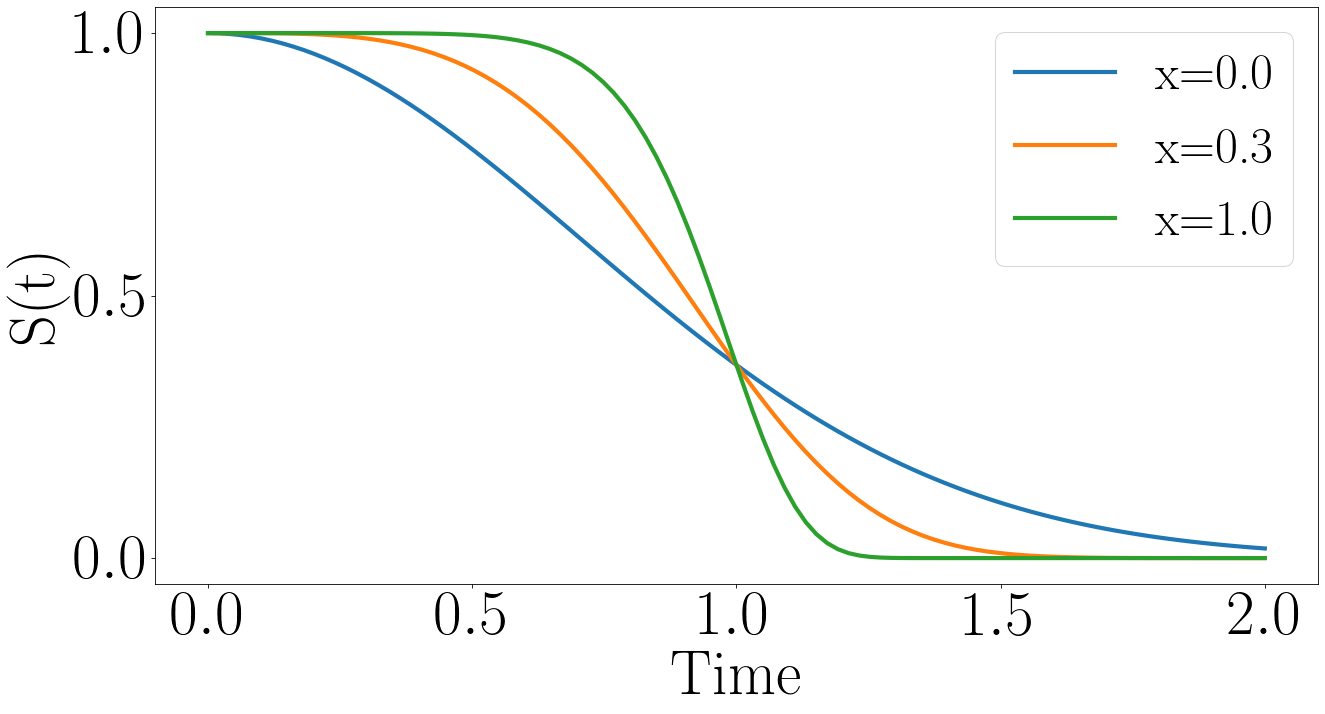}
    \end{subfigure}%
        \begin{subfigure}[t]{0.33\textwidth}
        \includegraphics[width=\linewidth]{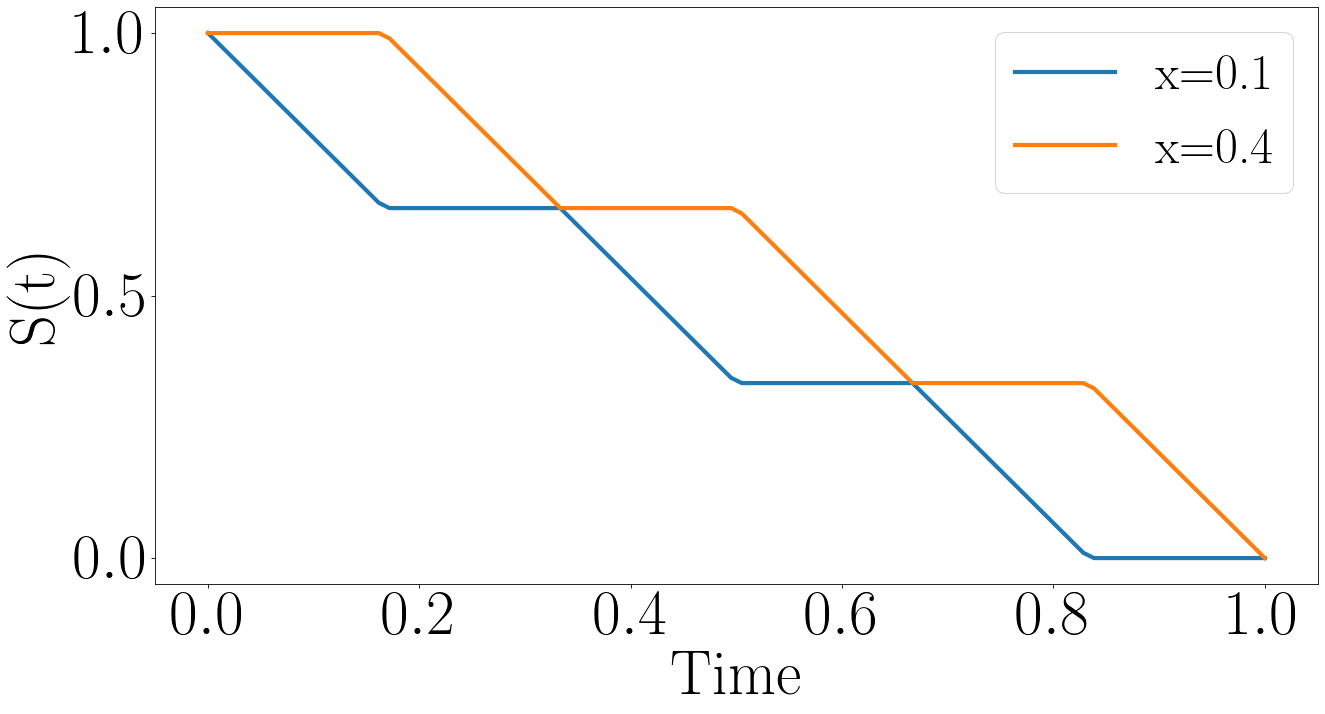}
    \end{subfigure}%
        \begin{subfigure}[t]{0.33\textwidth}
        \includegraphics[width=\linewidth]{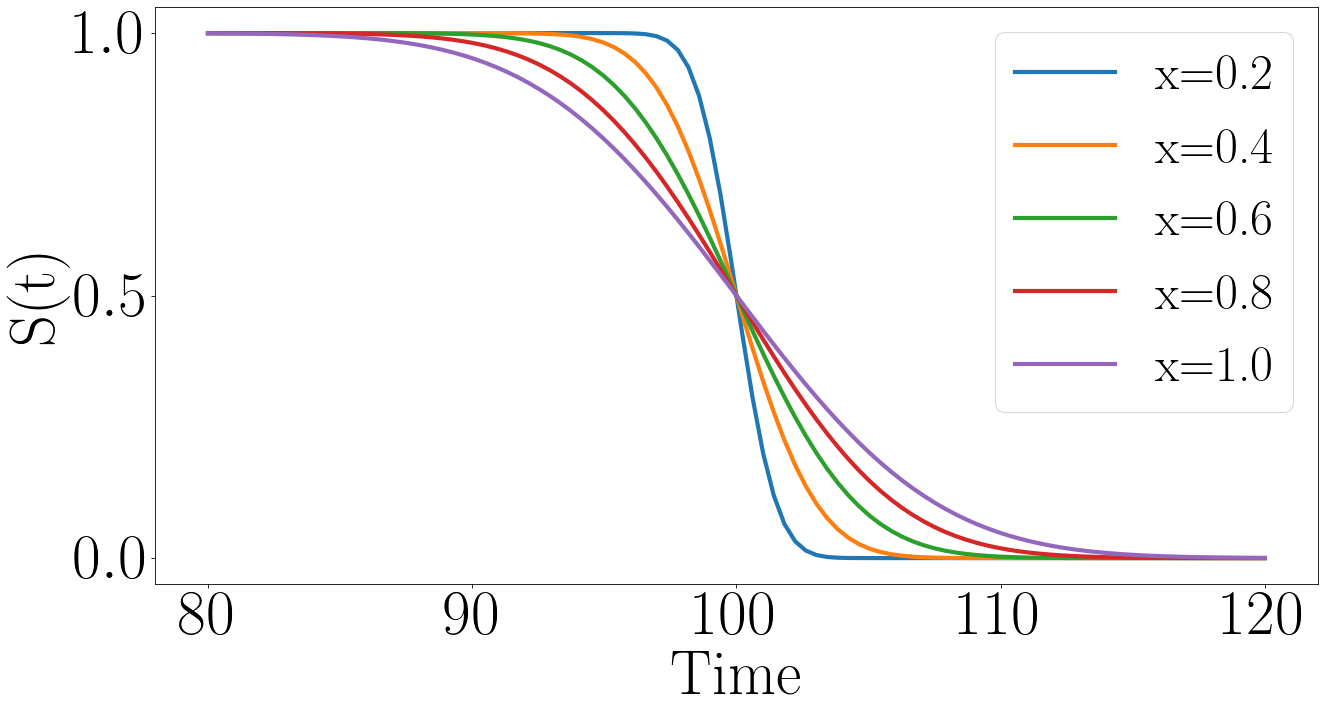}
    \end{subfigure}
    \end{subfigure}
        \begin{subfigure}[t]{\textwidth}
    \begin{subfigure}[t]{0.33\textwidth}
        \includegraphics[width=\linewidth]{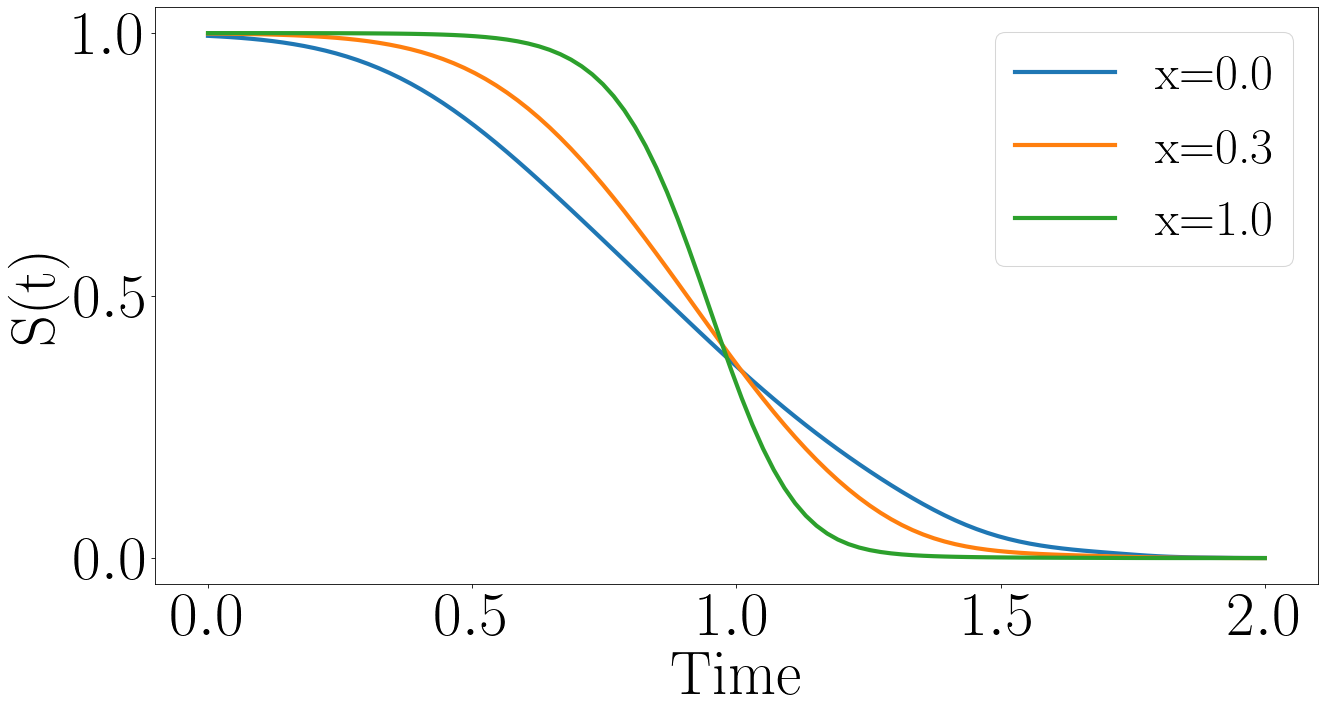}
    \end{subfigure}%
        \begin{subfigure}[t]{0.33\textwidth}
        \includegraphics[width=\linewidth]{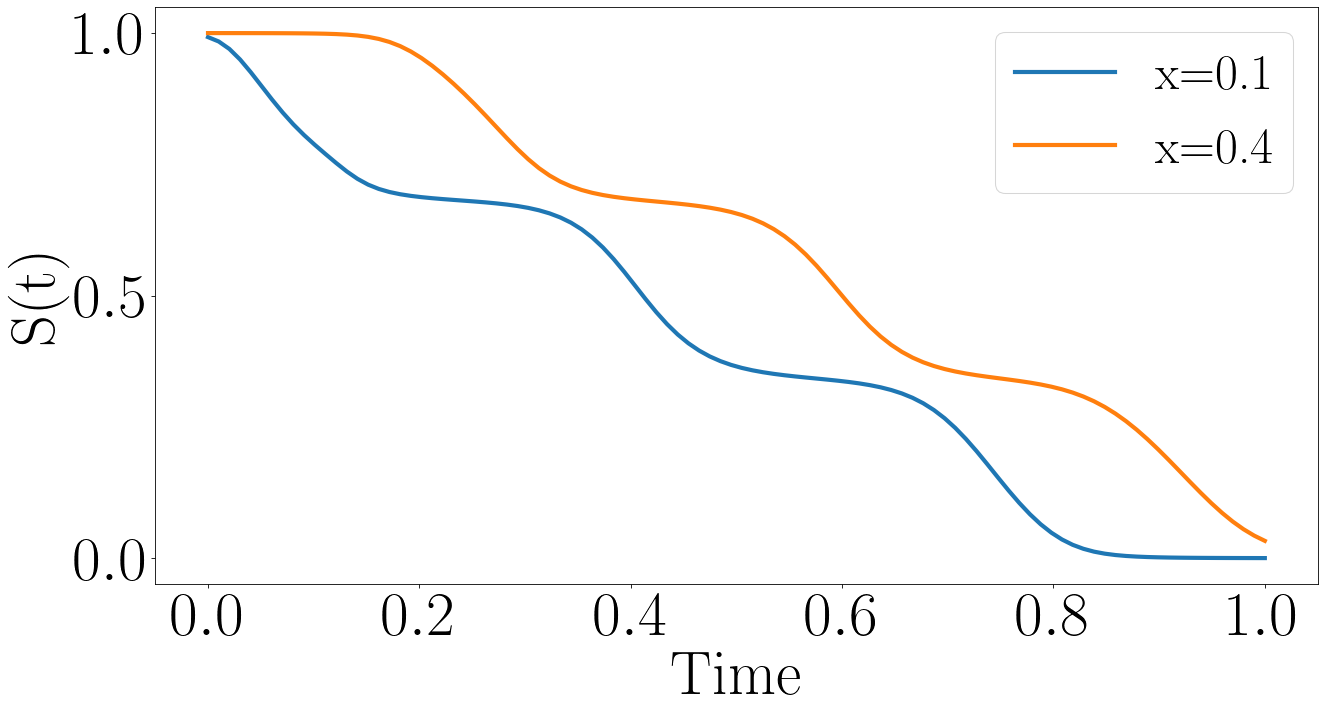}
    \end{subfigure}%
        \begin{subfigure}[t]{0.33\textwidth}
        \includegraphics[width=\linewidth]{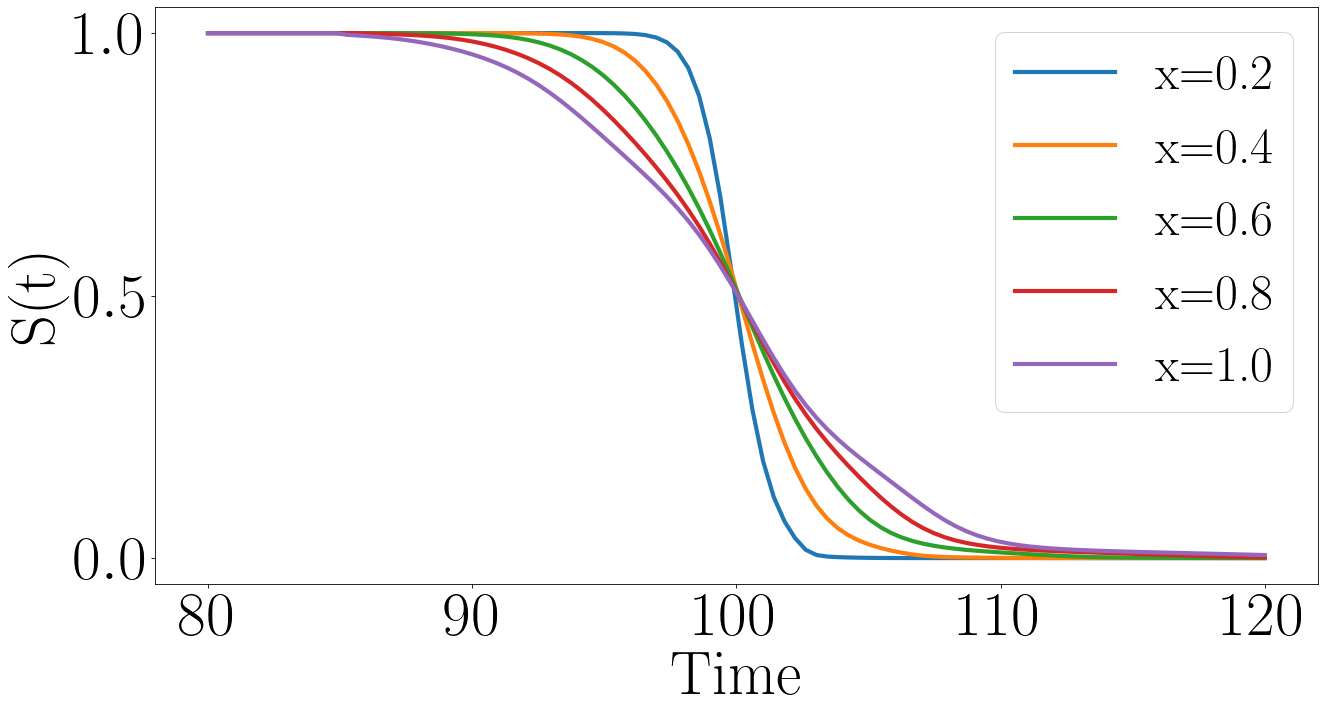}
    \end{subfigure}

    \end{subfigure}
\vspace{-0.4cm}
    \caption{True distributions (top row) and learned distributions using SuMo-net (bottom row). From left to right, the distributions are Weibull, Checkboard, and Normal.}
    \label{figure:toy_data}
    \vspace{-0.4cm}
\end{figure*}

\textbf{Results} \quad In Table \ref{table:likelihood} we compare the methods in terms of right-censored likelihood. As the benchmarks do not have tractable right-censored likelihood, this is approximated by interpolating their survival curves. While the obtained likelihoods of the benchmarks are only approximations, the results suggest SuMo-net is performant in terms of right-censored likelihood even when compared to much more complex NeuralODE based methods. It should be noted that neither SurvNODE or SODEN has been run on datasets larger than 40000 observations, due to computational costs scaling with observations and function evaluations \citep{massaroli2021dissecting}. In Table \ref{table:ibll} we compare the binomial log-likelihood of the models. Also in terms of IBLL, SuMo-net scores well. Results for IBS and time-dependent concordance are given in the Appendix. Notably, we find that DeepHit, which optimizes a concordance-like loss function, generally has the best performance in terms of time-dependent concordance, yet achieves the worst scores in terms of likelihood. This agrees with our findings that time-dependent concordance is not a proper scoring rule and that optimizing time-dependent concordance may lead to inaccurate distributions. 

\begin{table*}
\resizebox{\linewidth}{!}{%
\begin{tabular}{llllllllll}
\hline
\multicolumn{10}{c}{$S_L$ $\uparrow$}                                                                                                                                                               \\
 & CoxCC            & Cox-Linear       & Cox-Time         & DeepHit          & DeepSurv         & SODEN             & Weibull          & Log-normal       & SuMo-net                  \\ \hline
FLCHAIN        & $-0.507\pm0.014$ & $-0.513\pm0.014$ & $-0.584\pm0.239$ & $-0.85\pm0.493$  & $-0.529\pm0.015$ & N/A               & $-0.467\pm0.085$ & $-0.535\pm0.251$ & $\textbf{-0.376}\pm0.008$ \\
GBSG           & $-0.501\pm0.038$ & $-0.498\pm0.04$  & $-0.454\pm0.034$ & $-0.719\pm0.19$  & $-0.502\pm0.04$  & N/A               & $-0.577\pm0.117$ & $-0.563\pm0.115$ & $-\textbf{0.391}\pm0.027$ \\
KKBOX          & $0.797\pm0.013$  & $0.706\pm0.01$   & $1.411\pm0.014$  & $-4.174\pm4.128$ & $0.758\pm0.036$  & N/A               & $-0.118\pm 0.0$  & $0.201\pm0.289$  & $\textbf{1.553}\pm0.132$  \\
METABRIC       & $-0.305\pm0.037$ & $-0.316\pm0.031$ & $-0.233\pm0.066$ & $-0.565\pm0.348$ & $-0.318\pm0.029$ & $-0.149\pm 0.015$ & $-0.29\pm0.042$  & $-0.299\pm0.051$ & $-\textbf{0.142}\pm0.042$ \\
SUPPORT        & $0.533\pm0.047$  & $0.545\pm0.03$   & $0.57\pm0.038$   & $0.098\pm0.141$  & $0.549\pm0.045$  & $-0.676\pm 0.009$ & $0.035\pm0.024$  & $0.527\pm0.098$  & $\textbf{0.658}\pm0.027$  \\ \hline
\end{tabular}
}
\vspace{-0.2cm}

        \caption{Log-likelihood scores and standard deviations over the 5 folds. Higher is better. SuMo-net, Weibull, Log-normal and SODEN provides exact likelihood. For other methods we have approximated the likelihood. In particular, we approximate the density by interpolating the survival curve. See \Cref{est_likelihood_equation} for details.} \label{table:likelihood}
        \vspace{-0.4cm}

\end{table*}

\begin{table*}
\resizebox{\linewidth}{!}{%
\begin{tabular}{lllllllllll}
\hline
\multicolumn{11}{c}{$\mathcal{S}_{\text{IBLL}}$ $\uparrow$}                                                                                                                                                                              \\
 & CoxCC            & Cox-Linear       & Cox-Time                 & DeepHit          & DeepSurv         & SurvNODE          & SODEN                      & Weibull          & Log-normal       & SuMo-net                  \\ \hline
FLCHAIN        & $-0.464\pm0.194$ & $-0.336\pm0.012$ & $-0.47\pm0.241$          & $-0.646\pm0.462$ & $-0.339\pm0.01$  & N/A               & N/A                        & $-0.389\pm0.045$ & $-0.504\pm0.209$ & $-\textbf{0.332}\pm0.006$ \\
GBSG           & $-0.546\pm0.02$  & $-0.549\pm0.011$ & $-0.541\pm0.021$         & $-1.157\pm1.287$ & $-0.539\pm0.013$ & N/A               & N/A                        & $-1.513\pm1.525$ & $-1.093\pm0.633$ & $-\textbf{0.528}\pm0.012$ \\
KKBOX          & $-0.382\pm0.005$ & $-0.429\pm0.001$ & $-\textbf{0.33}\pm0.004$ & $-0.745\pm0.513$ & $-0.368\pm0.002$ & N/A               & N/A                        & $-3.032\pm1.766$ & $-1.776\pm2.187$ & $-0.395\pm0.029$          \\
METABRIC       & $-0.53\pm0.035$  & $-0.522\pm0.012$ & $-0.526\pm0.012$         & $-0.656\pm0.156$ & $-0.52\pm0.013$  & $-\textbf{0.477}$ & $-0.484\pm 0.010$          & $-0.554\pm0.004$ & $-0.597\pm0.064$ & $-0.485\pm0.015$          \\
SUPPORT        & $-0.572\pm0.008$ & $-0.577\pm0.005$ & $-0.577\pm0.009$         & $-0.701\pm0.065$ & $-0.569\pm0.01$  & $-0.580$          & $-\textbf{0.561}\pm 0.002$ & $-0.705\pm0.013$ & $-0.654\pm0.053$ & $-0.569\pm0.004$          \\ \hline
\end{tabular}

}
\vspace{-0.2cm}
        \caption{IBLL scores and standard deviations over the 5 folds. Higher is better}\label{table:ibll}
        \vspace{-0.4cm}
\end{table*}

\textbf{Correlation between likelihood and existing metrics} \quad
Following our result in Section 3 that likelihood is the only proper score among those considered, we investigate how the other scores compare to likelihood by calculating the rank correlation between all scores. In particular, for each of the datasets FLCHAIN, GBSG, METABRIC, SUPPORT and KKBOX and each hyperparameter run of SuMo-net, we compute all scores on the test set and turn them into ranks. The correlation matrix averaged over the datasets is displayed in \Cref{correlation_matrix}. 
\begin{table}
    \centering
    \resizebox{0.6\linewidth}{!}{%

\begin{tabular}{lrrrr}
\toprule
{} &  $\mathcal{S}_L$ &  $\mathcal{S}_{C^{\text{td}}}$ &       $\mathcal{S}_{\text{IBS}}$  &       $\mathcal{S}_{\text{IBLL}}$ \\
\midrule
$\mathcal{S}_L$  &    1.000 &     0.174 &  0.688 &  0.851 \\
$\mathcal{S}_{C^{\text{td}}}$ &    0.174 &     1.000 &  0.223 &  0.179 \\
$\mathcal{S}_{\text{IBS}}$        &    0.688 &     0.223 &  1.000 &  0.841 \\
$\mathcal{S}_{\text{IBLL}}$        &    0.851&     0.179&  0.841 &  1.000 \\
\bottomrule
\end{tabular}

}
\vspace{-0.2cm}
    \caption{Rank correlation matrix between likelihood, concordance, IBS and IBLL. The correlation matrix is averaged over all datasets. }
    \label{correlation_matrix}
    \vspace{-0.4cm}
\end{table}
The experiment indicates that concordance is almost independent of likelihood, while IBLL exhibits the strongest correlation, suggesting that it could serve as a good proxy when the likelihood is not available.

\textbf{Estimation of the likelihood for models other than SuMo-net} \quad All the baseline models proposed in \cite{kvamme2019time}, learn discrete distributions and hence have no density and the likelihood cannot be computed. For those methods we use interpolation of the survival curve to approximate the density. Let the survival curve of the discrete distribution be denoted by $S$ and denote the points at which the survival curve makes jumps by $t_0\leq\dots\leq t_m$. Then, given a time $t$, we find $i$ so that $t_i\leq z < t_{i+1}$ and then estimate
\begin{align}
\label{est_likelihood_equation}
\hat f(t) = -\frac{S(t_{i+k})-S(t_{i-k+1})  }{t_{i+k}-t_{i-k+1}},
\end{align} 
where $k$ determines the width of the interval we use to approximate the derivative. We found that setting $k=1$ resulted in a highly variable density and low likelihood, and hence we use $k=2$ in the main text. SuMo-net has tractable likelihood, so this approximation does not have to be made. While comparing models with an approximated and exact likelihood score is not ideal, it illustrates a general issue with modern survival models: they are not endowed with a tractable likelihood, making it difficult to evaluate them on a proper score at test time. As likelihood is an easy proper score to calculate for survival models, we stress that future models should be developed with tractable likelihood evaluation procedure in mind.  \newline\\ 
\textbf{Calibration} \quad When training survival models on likelihood directly we have direct access to the density $f(t|x)$ \emph{and} to the cumulative distribution $F(t|x)$, which means we can quantify how calibrated the fitted survival distribution is. Following \cite{kuleshov2018accurate} and \citet{goldstein2021xcal}, a calibrated survival distribution would have an empirical frequency $\hat{p}_{j}=\frac{\left|\left\{z_{i} \mid F\left(z_{i}|x_i\right) \leq p_{j}, i=1, \ldots, n\right\}\right|}{n}$ equal to $p_j$. The calibration error is then defined as $\operatorname{cal}\left(F_{1}, y_{1}, \ldots, F_{T}, y_{T}\right)=\sum_{j=1}^{m} w_{j} \cdot\left(p_{j}-\hat{p}_{j}\right)^{2}$, where we take $w_j=1,\forall j$ and $p_j\in\{0.1,0.2,0.3,0.4,0.5,0.6,0.7,0.8,0.9\}$. As the empirical frequency can only be coarsely approximated without direct access to $F(t|x)$, we only evaluate calibration for methods that train directly on likelihood.

\begin{table}
        \centering
    \resizebox{\linewidth}{!}{%
    \begin{tabular}{llll}
\hline
\multicolumn{4}{c}{Calibration score}                                                  \\
 & Weibull         & Log-normal               & SuMo-net                 \\ \hline
FLCHAIN        & $1.846\pm0.641$ & $\textbf{0.934}\pm0.579$ & $1.165\pm0.116$          \\
GBSG           & $0.762\pm0.442$ & $0.635\pm0.245$          & $\textbf{0.128}\pm0.03$  \\
KKBOX          & $0.897\pm0.401$ & $0.596\pm0.443$          & $\textbf{0.112}\pm0.077$ \\
METABRIC       & $0.327\pm0.177$ & $0.561\pm0.517$          & $\textbf{0.097}\pm0.056$ \\
SUPPORT        & $0.342\pm0.276$ & $\textbf{0.237}\pm0.139$ & $0.313\pm0.049$          \\ \hline
\end{tabular}
}
\vspace{-0.2cm}
    \caption{SuMo-net generally provides a calibrated survival distribution}
    \label{tab:my_label}
    \vspace{-0.4cm}
\end{table}
\textbf{Inference and training timings} \quad Here we contrast the timing of predictions of the survival probability $S(t\vert x)$ using a trained model. We compare the time of SuMo-net against the time of Cox-Time presented in \cite{kvamme2019time} and SODEN \citep{tang2020soden}. To compute the survival probability, Cox-Time integrates the learned hazard function over time, which is computationally costly. In contrast, SuMo-net evaluates $S$ through a single forward pass. To compare the timings we initialize SuMo-net, Cox-Time and SODEN using an identical architecture and compare inference timings on the test-fold of all datasets in \Cref{table:computational_time}. The findings indicate SuMo-net does indeed achieve its anticipated speedup.

\begin{table}
    \centering
\resizebox{\linewidth}{!}{%
\begin{tabular}{llllll}
\hline
\multirow{2}{*}{} & \multirow{2}{*}{SODEN (s)} & \multirow{2}{*}{Cox-Time (s)} & \multirow{2}{*}{SuMo-net (s)} & \multicolumn{2}{c}{Speedup vs} \\
                  &                        &                           &                           & SODEN       & Cox-Time       \\ \hline
SUPPORT           & $1.124\pm0.070$        & $2.97\pm0.126$            & $0.047\pm0.004$           & $24$           & $63$              \\
METABRIC          & $0.459\pm0.044$        & $1.573\pm0.126$           & $0.016\pm0.001$           & $29$           & $100$             \\
GBSG              & N/A                    & $1.46\pm0.112$            & $0.021\pm0.002$           & N/A            & $69$              \\
FLCHAIN           & N/A                    & $4.199\pm0.246$           & $0.044\pm0.003$           & N/A            & $96$              \\
KKBOX             & N/A                    & $140.505\pm7.976$         & $7.034\pm1.092$           & N/A            & $20$              \\ \hline
\end{tabular}
}
\vspace{-0.2cm}
\caption{Inference timings on 20\% of the data. SuMo-net achieves a speedup factor of between 20 and 100.}
\label{table:computational_time}
\vspace{-0.4cm}
\end{table}

\vspace{-0.2cm}
\section{CONCLUSION}
\vspace{-0.1cm}
We have shown that several existing scoring rules for right-censored regression are not proper, i.e., they may not faithfully reflect the accuracy of a fitted distribution. We also proved that the right-censored log-likelihood is a proper scoring rule. While other scoring rules can still be useful for certain purposes, their limitations should be taken into consideration. Since the right-censored log-likelihood is thus attractive both as a scoring rule and to assess model fit, we proposed SuMo-net, a simple, scalable and flexible regression model for survival data, for which the right-censored log-likelihood is tractable by construction. Future fruitful directions would be to extend SuMo-net to more involved architectures and datatypes such as genetic data, image data and other high dimensional data.

\textbf{Downstream applications}\quad While SuMo-net considers a monotonic neural network, the idea of enforcing monotonicity on the time-to-event variable $t$ can be extended to any model with trainable weights to train directly on any type of censored survival likelihood. This allows fields such as epidemiology, medical statistics, and economics, among many others, to formulate highly flexible non-parametric models that can evaluate directly on a proper scoring rule instead of non-proper proxy objectives. This is particularly important in safety critical domains where accurately approximating the true underlying model impacts subsequent decision making. We hope that that the introduction of SuMo-net can inspire future developments of sophisticated survival models endowed with a tractable likelihood for proper evaluation and comparison between models.
\clearpage
\section*{ACKNOWLEDGMENTS}
We thank the reviewers for their helpful remarks. We are grateful for our colleagues Jean-Francois Ton, Rob Zinkov, Siu Lun Chau and Zoi Tsangalidou for their helpful comments and remarks. 

\bibliographystyle{plainnat}
\bibliography{refs}


\clearpage
\appendix

\thispagestyle{empty}

\onecolumn \makesupplementtitle
\section{Appendix}

\subsection{Details on time-dependent concordance example}

Let $X\in \{0,1\}$. Let $K\sim \text{Uniform}\{0, 2, 4, 6, 8 \}$ and let $U\sim \text{Uniform}[0,1]$. Then $T | X=0 \sim K + U$ and $T | X=1 \sim K + U + 1$. We further set $C\sim \text{Exponential}(\text{mean}=20)$. Note that the hazard rates of the two distributions are nonzero on disjoint sets. The False distribution in Figure 1, is constructed such that at each time, the distribution with the nonzero hazard rate has the lowest survival probability.

\subsection{Details on Brier score example}

Let $X \sim \text{Bernoulli}(0.1)$. We create an extreme example of unequal censoring in which $C \sim \text{Exp}(1)$ if $X=0$ and $C=\infty$ if $X=1$. Let $T\sim \text{Exponential}(10)$ (independent of $X$), so that $F_{\text{True}}(t) = 1- \exp(-t/10)$. Finally create a fake distribution $\hat F$ so that $\hat F(t\vert X=1) = F_{\text{true}}(t)$  and $\hat F(t \vert X=0)=1- \exp(-t/4)$. We take a sample of size $n=10^6$. We evaulate the Brier score at $t=4$. We evaluate the Integrated Brier score from $t=0$ to $t=40$. We repeat the procedure for BLL and IBLL.

\subsection{Discussion of survival-CRPS}

\subsubsection{Discussion of proof given in Avati et al} 
In \cite{gneiting2011comparing} it is stated that the following score is proper (see \cite{matheson1976scoring} for a proof), 

\begin{align*}
\mathcal S(F,z) & =    \int_0^{\infty} (F(r)-1\{r \geq z \})^2 u(r)dr
\\ & = \int_0^z F(r)^2 u(r)dz + \int_z^{\infty} (1-F(r))^2u(r)dr.
\end{align*}
for non-negative weight functions $u$. In \cite{avati2020countdown} it is argued that the survival-CRPS is simply the above score, with weight function equal to the uncensored region. It is not immediately clear how this is meant to be interpreted, as the weight function needs to be a non-random function. If it were a proper score when conditioned on a known censoring time $C=c$, for every value of $c$, then we could proceed as in Section 3.2, using the conditional independence of $C$ and $T$. Assuming $C=c$, the uncensored region equals $u(r)=1\{r\leq c \}$. 
Proceeding with this as the weight function we would obtain
\begin{align*}
S(F,z) & =    \int_0^{\infty} (F(r)-1\{r \geq z \})^2 1\{ r\leq c \}dr
\\ & = \int_0^z F(r)^2 1\{r\leq c \}dz + \int_z^{\infty} (1-F(r))^21\{r\leq c \}dr.
\end{align*}
Note that $z\leq c$ since $z=\min(t,c)$ and therefore the first indicator is always true and can be removed, to arrive at:

\begin{align*}
S(F,z) & =  \int_0^z F(r)^2 dr + \int_z^{\infty} (1-F(r))^21\{r\leq c \}dr
\\ &= \int_0^z F(r)^2 dr + \int_z^{c} (1-F(r))^2dr.
\end{align*}
Hence, the survival-CRPS has an additional term
\begin{align} \label{eqn:missing}
    d \int_c^{\infty} (1-F(r))^2 dr
\end{align}
which is missing from the above expression. This shows, at least, that this interpretation of the proof strategy would not succeed. Furthermore, assuming that $c$ is a fixed censoring time, then $F(r)$ is only evaluated for $r>c$ in Equation \ref{eqn:missing}. Clearly to minimize the survival-CRPS, one then sets $F(r)=1$ for $r\geq c$, even if the true distribution satisfies $F_{\text{true}}(r)<1$ for $r\geq c$. 

\subsubsection{An example}

Assume for simplicity that there are no covariates. Let $T\sim \text{Exp}(\text{mean}=100)$ with every individual set to fail at $t=200$ so that $F_{True}(t)=1-\exp(-t/100)$ for $t <200$ and $F_{\text{true}}(200)=1$. Let $C\sim \text{Exp}(\text{mean}=10)$. Observations are thus likely to be censored. Let a fake distribution be $F_{\text{False}}= \text{Exp}(\text{mean}=25)$. We then take a sample of size $10^6$ from the fake distribution and true distribution and evaluate and compute their survival-CRPS against the true distribution. 


\subsection{Calculating life expectancies using SuMo-net}

The life expectancies are calculated by direct numerical integration using the learned survival distribution $S_\theta(t|x)$ since $\mathbb{E}\left[T\right | x]= \int_{0}^{\infty}  S_\theta(t| x) dt $.

\subsection{Log-likelihood scores for $d=1,4,8$}

\begin{table}[H]
\resizebox{\linewidth}{!}{%
\begin{tabular}{llllll}
\toprule
{} & \multicolumn{5}{l}{$\mathcal{S}_L \uparrow$} \\
dataset &           FLCHAIN &              GBSG &             KKBOX &          METABRIC &          SUPPORT \\
Method                       &                   &                   &                   &                   &                  \\
\midrule
CoxCC     &  $-0.583\pm0.014$ &  $-0.632\pm0.032$ &   $0.156\pm0.075$ &   $-0.451\pm0.06$ &   $0.437\pm0.06$ \\
Cox-Linear &  $-0.586\pm0.026$ &  $-0.603\pm0.032$ &    $0.05\pm0.111$ &   $-0.444\pm0.03$ &  $0.451\pm0.053$ \\
Cox-Time   &  $-0.565\pm0.071$ &  $-0.569\pm0.063$ &    $0.926\pm0.09$ &  $-0.346\pm0.056$ &  $0.481\pm0.047$ \\
DeepHit    &  $-0.878\pm0.496$ &  $-0.805\pm0.212$ &  $-1.432\pm0.226$ &    $-0.686\pm0.5$ &   $0.04\pm0.159$ \\
DeepSurv   &   $-0.605\pm0.01$ &   $-0.609\pm0.04$ &   $0.179\pm0.073$ &  $-0.455\pm0.027$ &  $0.446\pm0.036$ \\
SuMo-net   & $-\textbf{0.376}\pm0.008$        & $-\textbf{0.391}\pm0.027$ & $\textbf{1.553}\pm0.132$ & $-\textbf{0.142}\pm0.042$  & $\textbf{0.658}\pm0.027$ \\ 
\bottomrule
\end{tabular}

}
\caption{Log-likelihood scores for $d$ = 1}

\end{table}

\begin{table}[H]
\resizebox{\linewidth}{!}{%
\begin{tabular}{llllll}
\hline
 & \multicolumn{5}{l}{$\mathcal{S}_L \uparrow$} \\
Dataset & FLCHAIN & GBSG & KKBOX & METABRIC & SUPPORT \\
Method &  &  &  &  &  \\ \hline
CoxCC & $-0.482\pm0.01$ & $-0.451\pm0.033$ & $0.181\pm0.088$ & $-0.286\pm0.054$ & $0.564\pm0.04$ \\
Cox-Linear & $-0.49\pm0.017$ & $-0.465\pm0.043$ & $0.095\pm0.113$ & $-0.265\pm0.02$ & $0.565\pm0.032$ \\
Cox-Time & $-0.496\pm0.166$ & $-0.409\pm0.043$ & $0.93\pm0.089$ & $-0.189\pm0.061$ & $0.6\pm0.041$ \\
DeepHit & $-0.846\pm0.497$ & $-0.94\pm0.415$ & $-4.454\pm4.441$ & $-0.56\pm0.338$ & $0.113\pm0.126$ \\
DeepSurv & $-0.498\pm0.015$ & $-0.457\pm0.021$ & $0.158\pm0.117$ & $-0.282\pm0.027$ & $0.562\pm0.029$ \\
SuMo-net   & $-\textbf{0.376}\pm0.008$        & $-\textbf{0.391}\pm0.027$ & $\textbf{1.553}\pm0.132$ & $-\textbf{0.142}\pm0.042$  & $\textbf{0.658}\pm0.027$ \\ \hline
\end{tabular}
}
\caption{Log-likelihood scores for $d = 4$}

\end{table}

\begin{table}[H]
\resizebox{\linewidth}{!}{%
\begin{tabular}{llllll}
\hline
 & \multicolumn{5}{l}{$\mathcal{S}_L \uparrow$} \\
Dataset & FLCHAIN & GBSG & KKBOX & METABRIC & SUPPORT \\
Method &  &  &  &  &  \\ \hline
CoxCC & $-0.485\pm0.014$ & $-0.451\pm0.033$ & $0.42\pm0.021$ & $-0.243\pm0.023$ & $0.568\pm0.036$ \\
Cox-Linear & $-0.477\pm0.012$ & $-0.442\pm0.033$ & $0.344\pm0.037$ & $-0.253\pm0.012$ & $0.572\pm0.037$ \\
Cox-Time & $-0.55\pm0.178$ & $-0.398\pm0.036$ & $0.985\pm0.037$ & $-0.179\pm0.064$ & $0.599\pm0.043$ \\
DeepHit & $-0.835\pm0.491$ & $-0.637\pm0.138$ & $-3.622\pm3.38$ & $-0.552\pm0.299$ & $0.086\pm0.134$ \\
DeepSurv & $-0.495\pm0.01$ & $-0.435\pm0.025$ & $0.414\pm0.04$ & $-0.249\pm0.018$ & $0.573\pm0.042$ \\
SuMo-net   & $-\textbf{0.376}\pm0.008$        & $-\textbf{0.391}\pm0.027$ & $\textbf{1.553}\pm0.132$ & $-\textbf{0.142}\pm0.042$  & $\textbf{0.658}\pm0.027$ \\ \hline
\end{tabular}
}
\caption{Log-likelihood scores for $d = 8$}
\end{table}

\subsection{Other experiments}
\begin{table}[H]
\resizebox{\linewidth}{!}{%
\begin{tabular}{llllll}
\toprule
{} & \multicolumn{5}{l}{$\mathcal{S}_{C^{\text{td}}}$ $\uparrow$} \\
Dataset &          FLCHAIN &             GBSG &            KKBOX &         METABRIC &          SUPPORT \\
Method                       &                  &                  &                  &                  &                  \\
\midrule
CoxCC     &  $0.792\pm0.005$ &  $0.672\pm0.014$ &  $0.832\pm0.001$ &   $0.642\pm0.03$ &  $0.608\pm0.011$ \\
Cox-Linear &  $0.792\pm0.006$ &   $0.66\pm0.014$ &  $0.795\pm0.001$ &  $0.633\pm0.022$ &  $0.596\pm0.012$ \\
Cox-Time   &  $0.792\pm0.005$ &  $0.678\pm0.017$ &  $\textbf{0.848}\pm0.008$ &  $0.649\pm0.026$ &  $0.613\pm0.007$ \\
DeepHit    &  $\textbf{0.796}\pm0.004$ &  $\textbf{0.683}\pm0.013$ &  $0.799\pm0.005$ &  $\textbf{0.692}\pm0.022$ &  $\textbf{0.642}\pm0.003$ \\
DeepSurv   &  $0.791\pm0.004$ &  $0.671\pm0.014$ &  $0.828\pm0.001$ &   $0.645\pm0.02$ &  $0.609\pm0.011$ \\
SuMo-net  &   $0.79\pm0.004$ &  $0.671\pm0.013$ &  $0.76\pm0.045$ &  $0.658\pm0.029$ &  $0.603\pm0.004$ \\
SODEN  &   N/A &  N/A &  N/A &  $0.661\pm0.004$ &  $0.624\pm0.004$ \\
SurvNODE  &   N/A &  N/A &  N/A &  $0.667$ &  $0.622$ \\
\bottomrule
\end{tabular}

}
        \caption{Concordance scores. Higher is better}
\end{table}

\begin{table}[H]
\resizebox{\linewidth}{!}{%
\begin{tabular}{llllll}
\toprule
{} & \multicolumn{5}{l}{$\mathcal{S}_{\text{IBS}} \downarrow$} \\
Dataset &          FLCHAIN &             GBSG &            KKBOX &         METABRIC &          SUPPORT \\
Method                       &                  &                  &                  &                  &                  \\
\midrule
CoxCC     &  $0.112\pm0.014$ &  $0.185\pm0.006$ &  $0.118\pm0.001$ &  $0.179\pm0.013$ &  $\textbf{0.194}\pm0.003$ \\
Cox-Linear &  $0.102\pm0.004$ &  $0.188\pm0.004$ &    $0.138\pm0.0$ &  $0.176\pm0.005$ &  $0.197\pm0.002$ \\
Cox-Time   &  $0.114\pm0.017$ &  $0.183\pm0.008$ &  $\textbf{0.104}\pm0.001$ &  $0.178\pm0.005$ &  $0.196\pm0.004$ \\
DeepHit    &  $0.162\pm0.079$ &  $0.227\pm0.064$ &   $0.16\pm0.012$ &   $0.21\pm0.037$ &  $0.227\pm0.019$ \\
DeepSurv   &  $0.103\pm0.004$ &  $0.183\pm0.005$ &  $0.116\pm0.001$ &  $0.176\pm0.005$ &  $0.194\pm0.004$ \\
SuMo-net  &  $\textbf{0.101}\pm0.004$ &  $\textbf{0.179}\pm0.005$ &  $0.13\pm0.023$ &  $0.163\pm0.006$ &  $0.196\pm0.003$ \\
SODEN  &   N/A &  N/A &  N/A &  $0.162\pm0.003$ &  $\textbf{0.190}\pm0.001$ \\
SurvNODE  &   N/A &  N/A &  N/A &  $\textbf{0.157}$ &  $0.198$ \\
\bottomrule
\end{tabular}

}
        \caption{IBS scores. Lower is better}
\end{table}

\begin{table*}[t]
\resizebox{\linewidth}{!}{%

\begin{tabular}{lllllll|llllll}
\hline
         & \multicolumn{6}{c|}{Inference Times}                                                                                                                                                   & \multicolumn{6}{c}{Training times (1 epoch)}                                                                                                                                          \\ \hline
         & \multicolumn{1}{c}{CoxCC} & \multicolumn{1}{c}{Cox linear} & \multicolumn{1}{c}{Cox-Time} & \multicolumn{1}{c}{Deephit} & \multicolumn{1}{c}{DeepSurv} & \multicolumn{1}{c|}{SuMo-net} & \multicolumn{1}{c}{CoxCC} & \multicolumn{1}{c}{Cox linear} & \multicolumn{1}{c}{Cox-Time} & \multicolumn{1}{c}{Deephit} & \multicolumn{1}{c}{DeepSurv} & \multicolumn{1}{c}{SuMo-net} \\ \hline
SUPPORT  & $0.027\pm0.003$           & $0.026\pm0.003$                & $3.769\pm0.056$              & $0.003\pm0.0$               & $0.023\pm0.001$              & $0.044\pm0.0$                 & $0.664\pm0.068$           & $4.565\pm6.892$                & $7.411\pm1.273$              & $0.347\pm0.136$             & $0.525\pm0.027$              & $3.236\pm0.026$              \\
METABRIC & $0.008\pm0.001$           & $0.008\pm0.0$                  & $1.586\pm0.003$              & $0.001\pm0.0$               & $0.008\pm0.0$                & $0.016\pm0.0$                 & $0.156\pm0.01$            & $0.153\pm0.023$                & $2.676\pm0.02$               & $0.136\pm0.002$             & $0.147\pm0.002$              & $0.337\pm0.01$               \\
GBSG     & $0.008\pm0.0$             & $0.008\pm0.0$                  & $1.737\pm0.029$              & $0.002\pm0.0$               & $0.009\pm0.001$              & $0.021\pm0.0$                 & $0.194\pm0.069$           & $0.161\pm0.028$                & $2.829\pm0.053$              & $0.147\pm0.003$             & $0.159\pm0.013$              & $0.395\pm0.01$               \\
FLCHAIN  & $0.021\pm0.0$             & $0.021\pm0.001$                & $4.794\pm0.007$              & $0.003\pm0.0$               & $0.021\pm0.0$                & $0.041\pm0.001$               & $0.403\pm0.027$           & $0.394\pm0.028$                & $7.997\pm0.019$              & $0.208\pm0.002$             & $0.386\pm0.014$              & $1.605\pm0.062$              \\
KKBOX    & $1.836\pm0.021$           & $1.804\pm0.002$                & $158.141\pm0.121$            & $0.351\pm0.048$             & $1.814\pm0.021$              & $5.806\pm0.009$               & $990.645\pm0.817$         & $990.144\pm1.737$              & $1156.948\pm0.338$           & $872.381\pm1.083$           & $987.592\pm1.365$            & $700.269\pm0.517$            \\ \hline
\end{tabular}}
\caption{Additional experiments detailining run-times: SuMo-net is the fastest in terms of training speed per epoch on larger datasets.}
\label{exp}
\vspace{-0.5cm}
\end{table*}

\subsection{Implementation Details}

We present the implementation details to SuMo-net, including network architecture parameters, optimizer parameters and the code base.

\subsubsection{Hyperparameter space}
\begin{table}[H]
    \centering
    \begin{tabular}{lr}
        \toprule
        Hyperparameter & Values\\
        \midrule
        Layers      &    \{1, 2,4,5\} \\
        Layers (Covariate part)     &    \{1, 2,4,5\} \\
        Nodes per layer & \{8,16,32\} \\
        Nodes per layer  (Covariate part) & \{8,16,32\} \\
        Dropout & [0.0,0.1,0.2,0.3,0.4,0.5] \\
        Weigh decay & \{0.4, 0.2, 0.1, 0.05, 0.02, 0.01, 0\} \\
        Batch size & \{5,10,25,50,100,250\} \\
        \bottomrule
    \end{tabular}
    \caption{Hyperparameter search space for experiments on  Rot.~\& GBSG, SUPPORT, METABRIC, and FLCHAIN.}\label{tab:hyper_pars}
\end{table}

\begin{table}[H]
    \centering
    \begin{tabular}{lr}
        \toprule
        Hyperparameter & Values\\
        \midrule
        Layers      &    \{1, 2,4\} \\
        Layers (Covariate part)     &    \{4, 6, 8\} \\
        Nodes per layer & \{8, 16, 32,64\} \\
        Nodes per layer  (Covariate part) & \{128, 256, 512\} \\
        Dropout & [0, 0.7] \\
        Weigh decay & \{0.4, 0.2, 0.1, 0.05, 0.02, 0.01, 0\} \\
        Batch size & \{1000,2500,5000\} \\
        \bottomrule
    \end{tabular}
    \caption{KKBox hyperparameter configurations. (*) denotes parameters found with a two layer network with 128 nodes.}\label{tab:hyper_pars_kkbox}
\end{table}

For exact details, we refer to the code base attached. 

\subsubsection{Toy datasets}

We provide here the distributions of the toy datasets. In each of the distributions $X\sim \text{Unif}[0,1]$.

\begin{table}[H]
    \centering
    \begin{tabular}{lll}
        \toprule
        {Dataset} & ${T\vert X}$ & ${C\vert X}$ \\ \hline
        Weibull  & $\text{Weib}(\text{shape}=2+6X)$ & $\text{Exp}(\text{mean}=1.5)$ \\
        Normal & $\text{N}(100,6X)$ & $\text{N}(100,6)$  \\
        Checkerboard & $\text{CB}(4,6\vert X)$ & $\text{Exp}(\text{mean}=1.5)$ \\
        \bottomrule
    \end{tabular}
    \caption{Distributions of the toy datasets Weibull, Normal and Checkerboard. $\text{Weib}(\text{shape}=s)$ refers to the Weibull distribution with shape $s$ and scale $1$. For the definition of $\text{CB}(4,6)$ see below. Parameters are chosen such that appropriate numbers of individuals is censored (see data description main text) and the survival curves of the Normal and Weibull distributions cross (see plots main text).} \label{tab:toy_data_distributions}
\end{table}
The distribution $\text{CB}(4,6\vert X)$ is defined as follows. Let $X$ be on the horizontal axis and let $T$ be on the vertical axis. We split the square $[0,1]^2$ up in a grid of $6\times 4$ equally large rectangles, where there are 4 columns and 6 rows. If $X$ is in the 1st or 3rd column, then $T$ is distributed uniformly over the 1st, 3rd, or 5th row. If $X$ is in the 2nd or 4th column, then $T$ is distributed uniformly over the 2nd, 4th or 6th row.

\end{document}